\documentclass[journal]{IEEEtran}
\usepackage{times}
\usepackage{epsfig}
\usepackage{graphicx}
\usepackage{amsmath}
\usepackage{amssymb}

\usepackage{xcolor}
\usepackage{soul}
\usepackage[utf8]{inputenc}
\usepackage[pagebackref=true,breaklinks=true,letterpaper=true,colorlinks,bookmarks=false]{hyperref}

\usepackage[figuresright]{rotating}
\usepackage{textcomp,subfigure,algorithm,algorithmic,multirow,upgreek,tabularx}
\usepackage{graphics,gensymb}
\usepackage{bm,cases,threeparttable}

\usepackage{setspace}

\graphicspath{{Figures/}}

\usepackage{amsfonts}
\def\R{\mathbb{R}} 
\usepackage{caption}
\usepackage{enumitem}
\setitemize{noitemsep,topsep=0pt,parsep=0pt,partopsep=0pt}
\ifCLASSINFOpdf
\else
\fi
\begin{document}
	

\title{When Dictionary Learning Meets Deep Learning: \\ Deep Dictionary Learning and Coding Network for \\ Image Recognition with Limited Data}
%
%
%

\author{Hao Tang,
	Hong Liu,
	Wei Xiao and 
	Nicu Sebe,~\IEEEmembership{Senior Member, IEEE}
\thanks{Hao Tang and Nicu Sebe are with the Department of Information Engineering and Computer Science (DISI), University of Trento, Trento 38123,~Italy. E-mail: hao.tang@unitn.it, sebe@disi.unitn.it.
}
\thanks{Hong Liu is with the Shenzhen Graduate School, Peking University, Shenzhen 518055, China. 
E-mail: hongliu@pku.edu.cn.
}
\thanks{Wei Xiao is with the Shenzhen Lingxi Artificial Intelligence Co., Ltd, Shenzhen 518109, China.
E-mail: xiaoweithu@163.com.
}
\thanks{Corresponding author: Hao Tang and Hong Liu.}}

%
%

\markboth{IEEE Transactions on Neural Networks and Learning Systems}%
{Shell \MakeLowercase{\textit{et al.}}: Bare Demo of IEEEtran.cls for IEEE Journals}
%



\maketitle



\begin{abstract}
We present a new Deep Dictionary Learning and Coding Network (DDLCN) for image recognition tasks with limited data.
The proposed DDLCN has most of the standard deep learning layers (e.g., input/output, pooling, fully connected,  etc.), but the fundamental convolutional layers are replaced by our proposed compound dictionary learning and coding layers.  
The dictionary learning learns an over-complete dictionary for input training data.
At the deep coding layer, a locality constraint is added to guarantee that the activated dictionary bases are close to each other. 
Then the activated dictionary atoms are assembled and passed to the compound dictionary learning and coding layers.
In this way, the activated atoms in the first layer can be represented by the deeper atoms in the second dictionary.
Intuitively, the second dictionary is designed to learn the fine-grained components shared among the input dictionary atoms, thus a more informative and discriminative low-level representation of the dictionary atoms can be obtained. 
We empirically compare DDLCN with several leading dictionary learning methods and deep learning models. 
Experimental results on five popular datasets show that DDLCN achieves competitive results compared with state-of-the-art methods when the training data is limited.
Code is available at \url{https://github.com/Ha0Tang/DDLCN}.
\end{abstract}


\section{Introduction}
\IEEEPARstart{T}{he} key step of classifying images is obtaining feature representations encoding relevant label information.
In the last decade, the most popular representation learning methods are dictionary learning (or sparse representation) and deep learning.
Dictionary learning is learning a set of atoms so that a given image can be well approximated by a sparse linear combination of these learned atoms, while deep learning methods aim at extracting deep semantic feature representations via a deep network. 
Scholars from various research fields have realized and promoted the progress of dictionary learning with great efforts, e.g.,~\cite{tibshirani1996regression, donoho2006compressed} from the statistics and machine learning community, \cite{engan1999method} and \cite{aharon2006k} from the signal processing community and~\cite{liu2016sequential,wright2009robust,tang2016novel} from the computer vision and image processing communities. However, what is a sparse representation and how can we benefit from it?
These two questions represent the points we attempt to clarify among the fundamental philosophies of sparse representation.
In the following, we start with a brief description of the sparse representation and then put more emphasis on the relationship between sparsity and locality, to elicit the research focus points of the paper.

\begin{figure}[!t] \small
	\centering
	\includegraphics[width=1\linewidth]{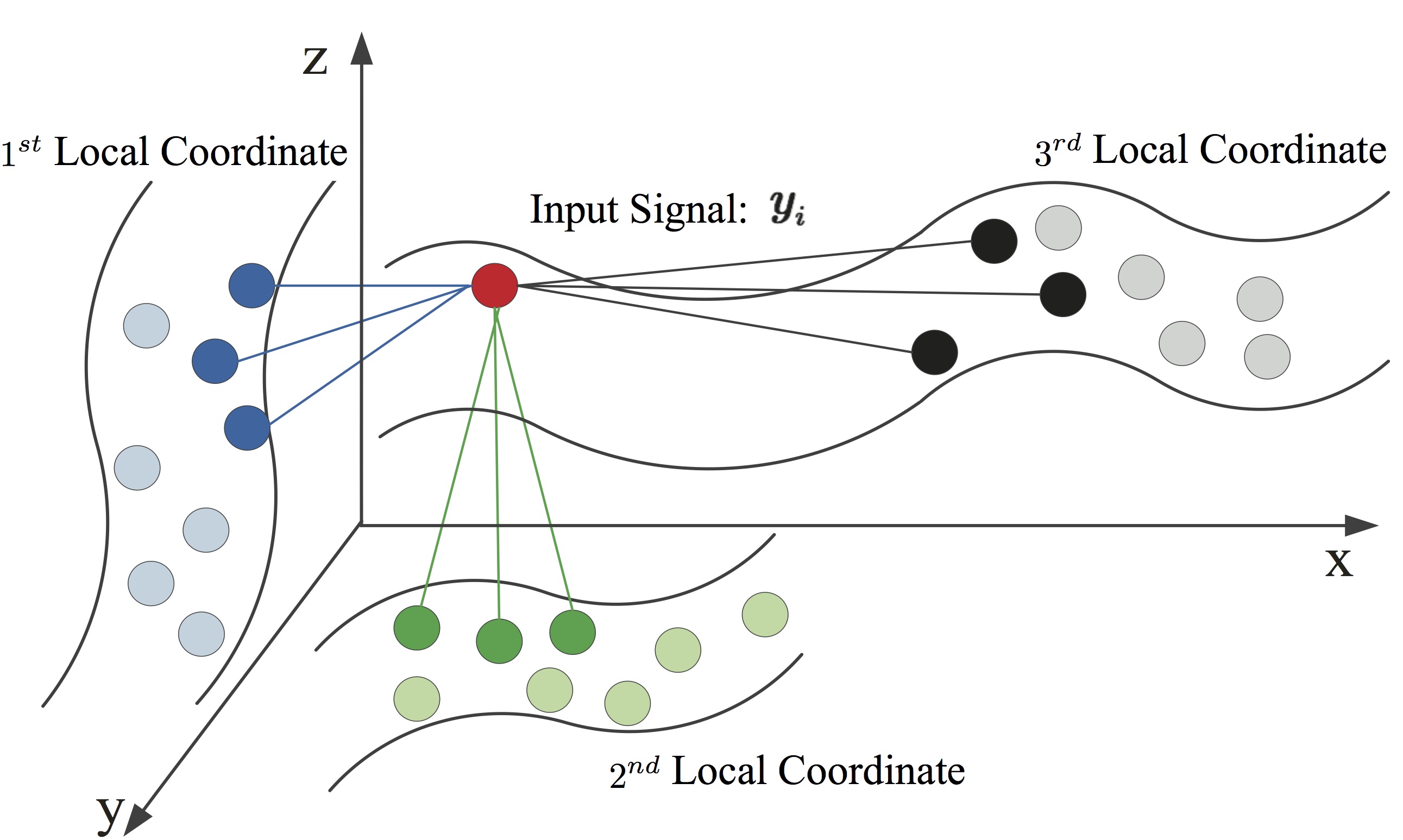}
	\caption{Multiple local coordinates and `fake' anchor points.
	}
	\label{manifold}
	\vspace{-0.4cm}
\end{figure}

\begin{figure*}[!tbp] \small
	\centering
	\includegraphics[width=1\linewidth]{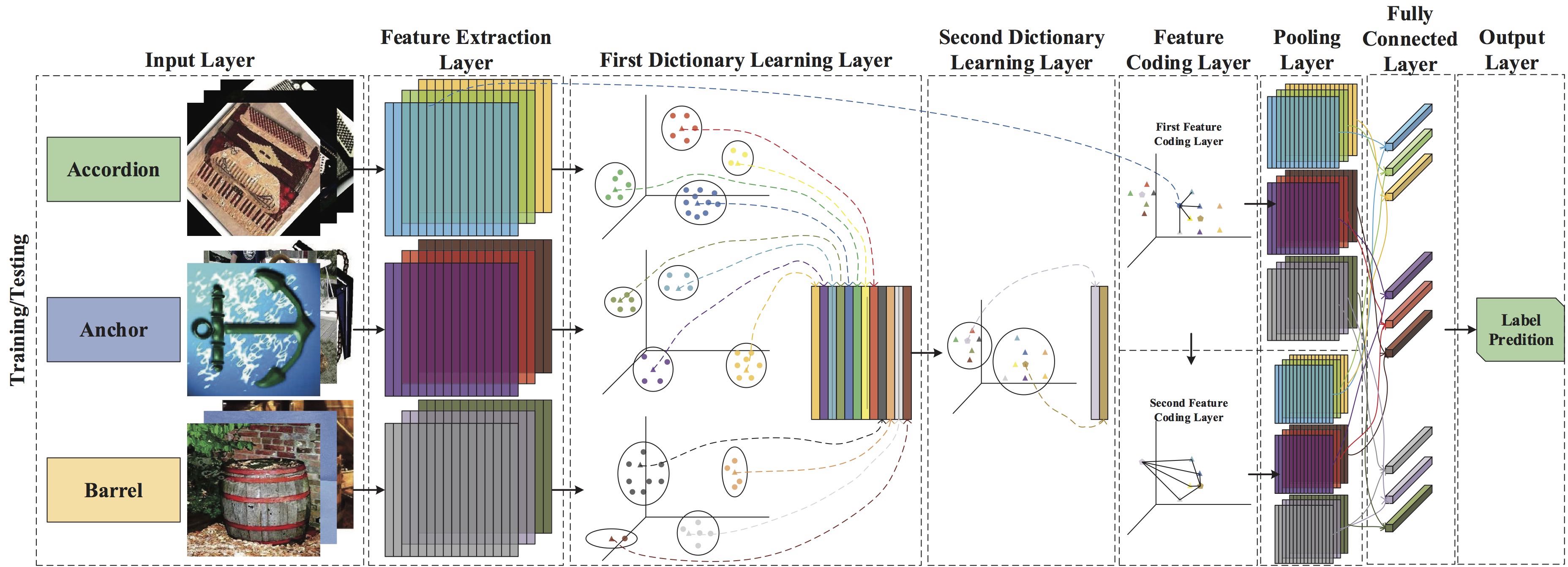}
	\caption{The framework of the proposed Deep Dictionary Learning and Coding Network (DDLCN).
	}
	\label{fig:framework}
	\vspace{-0.4cm}
\end{figure*}

\noindent \textbf{Dictionary Learning} \cite{donoho2006compressed, wright2009robust, mairal2010sparse}, also called sparse representation or sparse coding, represents a signal with a linear combination of a few elements/atoms from a dictionary matrix containing several prototype signal-atoms.
The coefficient vector specifies how we select those seemingly `useful' atoms from dictionary to linearly combine the original
signal, i.e., each entry of the coefficient vector corresponds to a specific atom, and its value is viewed as a weight that specifies the linear combination proportion of each selected atom.
The important asset of sparse representation is that the complex signal can be represented in a concise manner enabling the following classification procedure to adopt a simpler classifier (e.g., linear classifier).
This is a fundamental advantage of sparse coding.
Besides, it is worth noting that sparse-inducing regularization and sparse representation have been demonstrated to be extremely powerful for representing complex signals in recent works. Signals such as audios, videos and images admit naturally sparse representations, while the key idea of sparse representation is mapping the original chaotic signals to their corresponding concise representations with a regularized or uniform style. These sparse representations have an intimate relationship due to their over-complete dictionaries.

\noindent \textbf{Locality and Sparsity.}
Recently, Yu et al. \cite{yu2009nonlinear} extend the form of sparse representation and propose the Local Coordinate Coding (LCC) framework.
LCC points out that locality is an intrinsic property rather than the sparsity under certain assumptions since locality will lead to sparsity but not necessarily vice versa.
Specifically, LCC learns a non-linear high dimension function by forming an adaptive set of basis functions on the data manifold.
To achieve a higher approximation accuracy, coding should be as local as possible to be better incorporated by the subspace spanned by the adaptive set of basis functions. Thereafter, Wang et al. \cite{wang2010locality} further developed a fast version of LCC, i.e., Locality-constrained Linear Coding (LLC). LLC employs the locality constraint to project each feature into its local-coordinate system.
This fast version relaxes the local regularization term using a $\ell_2$ norm-based term. That is why it runs faster.
As with LLC, two modified versions are proposed in~\cite{yu2010improved} and~\cite{lin2010deep}, respectively, where a second layer is introduced to further improve the approximation power using different strategies, such as~\cite{yu2010improved}~with local PCA and~\cite{lin2010deep}~with deep coding network.
In the context of nonlinear function approximation, sparse coding brings an in-depth understanding of its fundamental connotation and success.
More importantly, it also obtains a deeper insight into its parentage ties with the essence, i.e., locality.

\noindent \textbf{Limitations.}
Methods such as LCC \cite{yu2009nonlinear} and LLC \cite{wang2010locality} are based on the observation that the sparse representations tend to be `local'.
However, these methods such as LLC have a major disadvantage, i.e., to achieve higher approximation, we need to employ a large number of so-called `anchor points' for making a better linear approximation of the original signal.
Since LLC is a local linear approximation of a complex signal $\bm{y}_i$, which means the anchor points need to provide higher approximation power, making some of them not to be necessarily `real' local anchors on the manifold where the original signal $\bm{y}_i$ resides.
Therefore, the goal of the paper is to equip anchors with higher descriptive power to better approximate the input $\bm{y}_i$ for making more accurate inferences from it (see Fig.~\ref{manifold} for a better understanding). Moreover, we observe that most studies in dictionary learning so far adopt a shallow (single layer) network architecture.
For instance, existing leading dictionary learning approaches are  K-SVD \cite{aharon2006k}, Discriminative K-SVD (D-KSVD) \cite{zhang2010discriminative} and Label Consistent K-SVD (LC-KSVD) \cite{jiang2011learning}, which aim to decompose the input data into a dense basis and sparse coefficients.
However, shallow network architectures are difficult to fully extract the intrinsic properties of the input data. In our preliminary experiments, we observe that both limitations lead to very poor classification performance when the data are limited.

\noindent \textbf{Deep Dictionary Learning.}
To fix both limitations, recent works such as~ \cite{tariyal2016greedy,liu2018dictionary,chun2018convolutional,hu2018nonlinear,xiao2015two,zhang2017jointly,nguyen2015dash,dong2018learning,song2019multi,mahdizadehaghdam2019deep} have demonstrated that deeper network architectures can be built from dictionary learning methods. 
For example, Liu et al.~\cite{liu2018dictionary} present a new dictionary learning layer to replace both the conventional fully connected layer and the rectified linear unit in a deep learning framework for scene recognition tasks.
Song et al.~\cite{song2019multi} propose MDDL with a locality constraint for image classification tasks.
Mahdizadehaghdam et al.~\cite{mahdizadehaghdam2019deep} introduce a new model, which tries to learn a hierarchy of deep dictionaries for image classification tasks.
Different from conventional deep neural networks, deep dictionary learning methods usually first extract feature representations at a patch-level, and then rely on the patch-based learned dictionary to output a global sparse feature representation for the input data.

Our proposed method borrows some useful ideas from CNNs and differs from these multi-layer dictionary learning methods in two aspects.
First, our dictionary is directly learned from image features and then the learned dictionary acts as a candidate pool for learning the next layer dictionary. 
The learned dictionaries from different layers have connections while existing methods use a fixed dictionary in different layers, i.e., there is no message passing between the dictionaries of different layers.
Second, to represent an atom in the previous layer, our model picks out a few atoms in the next layer and linearly combine them. 
The activated atoms have a linear contribution in constructing the atom in the previous layer. 
This could incorporate more information into the next layer’s codes and alleviate the influence of incorrect atoms. 

\noindent \textbf{Contributions.} In this paper, we aim to improve the deep representation ability of dictionary learning.
To this end, we propose a novel Deep Dictionary Learning and Coding Network (DDLCN), which mainly consists of several layers, i.e., input, feature extraction, dictionary learning, feature coding, pooling, fully connected and output layer, as shown in Fig.~\ref{fig:framework}.

The design motivation of the proposed DDLCN is derived from both Convolutional Neural Networks (CNNs) and dictionary learning approaches.
However, the biggest difference being that the convolutional layers in CNNs are replaced by our proposed dictionary learning and coding layers.
By doing so, the proposed DDLCN can learn edge, line and corner representations from the shallow dictionary layers. 
Then additional sophisticated `hierarchical' feature representations can be learned from deeper dictionary layers.
The proposed DDLCN has a better approximation capability of the input since the introduction of the proposed dictionary learning and coding layer, which takes advantage of the manifold geometric structure to locally embed points from the underlying data manifold into a lower-dimensional deep structural space.
Moreover, it also fully considers each fundamental basis vector adopted in the shallow layer coding,
and incorporates additional gradient effects of nonlinear functions on it into the deeper local representation.
Thus,  the proposed DDLCN can transfer a very difficult nonlinear learning problem into a simpler linear learning one.
More importantly, the approximation power is higher than its single-layer counterpart.

Overall, the contributions of the paper are summarized as: 
1) We present a novel deep dictionary learning DDLCN framework, which combines the advantages of both dictionary learning  and deep learning methods.
2) We introduce a novel compound dictionary learning and coding layer, which can substitute the convolutional layer in the standard deep learning architectures.
3) Extensive experiments on a broader range of datasets with limited training data demonstrate that the proposed method outperforms state-of-the-art dictionary learning approaches  and achieves competitive results compared with existing deep learning methods.

Part of this work has been published in \cite{tang2018deep}. The additional contributions are:
1) We present a more detailed analysis by including recently published works about deep dictionary learning.
2) We generalize the two-layer framework of DDLCN in \cite{tang2018deep} to a deeper one and validate the effectiveness.
3) We present an in-depth description of the proposed layer and framework, providing all the architectural and implementation details of the method.
4) We also extend the experimental evaluation provided in \cite{tang2018deep} in several directions. 
First, we conduct extensive experiments on five popular datasets, demonstrating the wide application scope of our DDLCN framework. 
Second, we conduct more interpretative experiments to show  the superiority of the proposed DDLCN compared with the traditional convolutional layers. Third, we introduce several variants of the proposed DDLCN, i.e., DDLCN-3, DDLCN-4, DDLCN-5 and DDLCN-6, which achieve better results than DDLCN-2 proposed in \cite{tang2018deep}. Lastly, we have also included new state-of-the-art baselines, e.g., MDDL \cite{song2019multi}, and we observe that the proposed DDLCN achieves always better results.

\section{Related Work}
\label{sec_rel}
In this section, we briefly review the related work.
Some important notations used in our paper are given in Table~\ref{table_natations}.

\begin{table}[!tbp] \small
	\caption{The notations used in this paper.}
	\centering
	\resizebox{\linewidth}{!}{
		\begin{tabular}{|cl|} \hline
			$\bm{Y}$ & $[{\bm{y}_1},{\bm{y}_2}, \cdots ,{\bm{y}_n}] = \{ {\bm{y}_i}\} _{i = 1}^n$ \\
			$\bm{y}$ & signal (also called data vector or descriptor or feature) \\
			$\bm{y'}$ & single-layer physical approximation of $\bm{y}$ \\
			$\bm{y''}$ & two-layer physical approximation of $\bm{y}$ \\
			$\bm{X}$ & $[{\bm{x}_1},{\bm{x}_2}, \cdots ,{\bm{x}_n}] = \{ {\bm{x}_i}\} _{i = 1}^n$ \\
			$\bm{x}$ & coefficient vector or solution \\
			${\bm{v}}$ & anchor point of ${\bm{C^1}}$ or first layer atom \\
			${\bm{v'}}$ & physical approximation of ${\bm{v}}$ \\
			${\bm{u}}$ & anchor point of $\bm{{C}^{2,v}}$ or second layer atom \\
			${\bm{C^1}}$ &  set of anchor points to $\bm{y}$ \\
			${\bm{C^{2,v}}}$ & set of anchor points to $\bm{v}$ \\	
			$\bm{{\gamma}^1}$ & map of $\bm{y}$ to $\bm{{\gamma}^1}(\bm{y})$ \\
			$\bm{{\gamma}^{2,v}}$ & map of $\bm{v}$ to $\bm{{\gamma}^{2,v}}(\bm{v})$ \\
			$\bm{D}$ & $[{\bm{d}_1},{\bm{d}_2}, \cdots ,{\bm{d}_p}]$, a dictionary or codebook with $\bm{p}$ atoms \\
			$\bm{d}$ & element or atom or codeword \\
			$D_i$ & dictionary size of the $\bm{i}$ layer   \\
			$\bm{m}$ & dimension of $\bm{y}$ \\
			$\bm{l}$ & number of signals \\
			$\bm{n}$ & number of deep layers \\
			\hline
		\end{tabular}}
	\vspace{-0.4cm}
	\label{table_natations}
\end{table}

Sparse coding (or sparse representation) represents an original signal~$\bm{y}$ with a sparse signal~$\bm{x}$ based on a dictionary~$\bm{D} {=} [{\bm{d}_1},{\bm{d}_2}, \cdots ,{\bm{d}_p}] {\in} \R^{m\times p}$, where the dictionary~$\bm{D}$ is learned by dictionary learning algorithms.
Usually, it is formulated as the following constrained optimization objective,
\begin{equation}
	\label{p0}
	\mathop {\min }\limits_{\bm{x}} {\left\| \bm{x} \right\|_0}\quad s.t.\quad \bm{y} = \bm{D}\bm{x},
\end{equation}
where $\left\|  \cdot  \right\|_0$ is the $\ell_0$ pseudo-norm, which aims to count the non-zeros entries of a vector.
However, the equality constraint $\bm{y} {=} \bm{D}\bm{x}$ is too strict for solving the problem.
Hence, we relax the optimization problem  with a small threshold as follows,
\begin{equation}
	\label{p0_epsilon}
	({P_{0,\varepsilon }})\quad \mathop {\min }\limits_{\bm{x}} {\left\| \bm{x} \right\|_0}\quad s.t.\quad {\left\| {\bm{y} - \bm{D}\bm{x}} \right\|_2} \le \varepsilon ,
\end{equation}
or equivalently, its corresponding unconstrained form is as follows using the Lagrange multipliers,
\begin{equation}
	\label{p0_lagrange}
	({P_{0,\lambda }})\quad \mathop {\min }\limits_{\bm{x}} \left[ {\frac{1}{2}\left\| {\bm{y} - \bm{D}\bm{x}} \right\|_2^2 + \lambda {{\left\| \bm{x} \right\|}_0}} \right].
\end{equation}

Unfortunately, the approaches that provide an approximate solution to the problem are pursuit procedures but are not global optimal because this problem is NP-hard.
In the past few decades, various pursuit procedures have been introduced.
For instance, both Matching Pursuit (MP) \cite{mallat1993matching} and Orthogonal Matching Pursuit (OMP) \cite{chen1989orthogonal} are greedy algorithms that select the dictionary atoms sequentially, which involves the computation of inner products between the input signal and dictionary atoms.

Another type of pursuit procedure is a relaxation strategy. 
For instance, Basis Pursuit (BP) \cite{chen2001atomic} converts the problems
of Eq.~\eqref{p0} or Eq.~\eqref{p0_epsilon} to their convex counterparts by replacing the $\ell_0$ pseudo-norm with the $\ell_1$ norm.
Focal Underdetermined System Solver (FOCUSS) \cite{gorodnitsky1997sparse} employs the $\ell_q$ norm with $q {\le} 1$ to replace the $\ell_0$ pseudo-norm.
Since the $\ell_0$ optimization problem is non-convex, one of the popular relaxations is Lasso~\cite{tibshirani1996regression}, which uses the $\ell_1$ norm instead of $\ell_0$ pseudo-norm,
\begin{equation}
	\label{lasso_form1}
	({P_{1,\varepsilon }})\quad \mathop {\min }\limits_{\bm{x}} {\left\| \bm{x} \right\|_1}\quad s.t.\quad {\left\| {\bm{y} - \bm{D}\bm{x}} \right\|_2} \le \varepsilon ,
\end{equation}
where $\left\|  \cdot  \right\|_1$ is the $\ell_1$ norm and the corresponding unconstrained form is as follows, using the Lagrange multipliers,

\begin{equation}
	\label{lasso_form2}
	({P_{1,\lambda }})\quad \mathop {\min }\limits_{\bm{x}} \left[ {\frac{1}{2}\left\| {\bm{y} - \bm{D}\bm{x}} \right\|_2^2 + \lambda {{\left\| \bm{x} \right\|}_1}} \right].
\end{equation}
It is well known that, as $\lambda$ goes larger, $\bm{x}$ tends to be more sparse, such that only a few dictionary elements are involved.

This paper focuses on learning the sparse representations in a situation where the data have only a few significant patterns. This greatly benefits the applications of classifications and information fusion.
There are several popular extensions of traditional sparse coding, i.e., Group Sparse Coding~\cite{yuan2006model, bach2008consistency}, LCC \cite{yu2009nonlinear, yu2010improved} and its fast implementation algorithm LLC \cite{wang2010locality}.
Group Sparse Coding encourages the solutions of sparse regularized problems to have the specific patterns of non-zero coefficients, which benefits higher-level tasks such as image recognition~\cite{roth2008group} and compressive sensing~\cite{huang2011learning}.
LCC and LLC empirically observe that sparse representation results tend to be `local'.
As indicated in LCC~\cite{yu2009nonlinear, wang2010locality}, the locality is more essential than sparsity, especially for supervised learning, since the locality will lead to sparsity but not necessarily vice versa.

The corresponding LLC \cite{wang2010locality} representations can be obtained by solving the convex programming,
\begin{equation}
\label{llc}
\mathop {\min }\limits_{\bm{x}} \left[ {\sum\limits_{i = 1}^n {\left\| {{\bm{y}_i} - \bm{D}{\bm{x}_i}} \right\|_2^2 + \lambda \left\| {{\bm{b}_i} \odot {\bm{x}_i}} \right\|_2^2} } \right] \\
\quad s.t.\quad {\bm{1}^{\mathsf{T}}}{\bm{x}_i} = 1,
\end{equation}
where $\bm{y}_i$ is the $i^{th}$ local descriptor of $\bm{Y} {=} \left[ {{\bm{y}_1}, \cdots ,{\bm{y}_n}} \right]$ and $\bm{X} {=} [\bm{x}_1, \bm{x}_2, \cdots, \bm{x}_n]$ is the set of codes for $\bm{Y}$. 
The constraint ${\bm{1}^{\mathsf{T}}}{\bm{x}_i} {=} 1$ follows the shift-invariant requirements of the LLC code. $\odot$ represents the element-wise multiplication.
${\bm{b}_i} {=} \exp (\frac{{dist({\bm{y}_i},\bm{D})}}{\psi})$ is the locality adapter, which provides different degrees of freedom to each basis vector $\bm{d}_j$ proportional to its similarity to the input descriptor $\bm{y}_i$,
where $dist({\bm{y}_i},\bm{D}) {=} {[dist({\bm{y}_i},{\bm{d}_1}), \cdots ,dist({\bm{y}_i},{\bm{d}_p})]^{\mathsf{T}}}$, $dist({\bm{y}_i},{\bm{d}_j})$ is the Euclidean distance
between $\bm{y}_i$ and $\bm{d}_j$, and $\psi$ is used for adjusting the weight decay speed for the locality adapter. 

The LLC solution in Eq.~\eqref{llc} is not sparse in the sense of $\ell_0$ pseudo-norm or any other sparse inducing norms, while it is really sparse in the sense that the solution
only has few non-zero values.
This idea enriches the connotation of `sparse', as it makes possible various applications such as signal representation with simultaneous sparse and discriminative properties.
This property of sparse is mainly induced by locality regularization term $\lambda \left\| {{\bm{b}_i} \odot {\bm{x}_i}} \right\|_2^2$.
More specifically, each entry in $\bm{b}_i$ constrains the corresponding entry in $\bm{x}_i$, i.e., the more non-zero entries in $\bm{b}_i$, the fewer values in the counterpart~$\bm{x}_i$.
This effect is amplified exponentially by~$\bm{b}_i$.

Compared with these methods, the newly introduced coding strategy captures more accurate correlations.
Traditional sparse coding methods only pursue the solo goal, i.e., to be as sparse as possible in the final representation.
While LLC aims to catch the delicate atoms structure of the manifold where the input signals reside, and it further uses these activated atoms for signal representation.
However, existing methods have a major limitation, i.e., to achieve higher approximation, one has to use a large number of so-called `anchor points' to achieve a better linear approximation of the input signal.
To fix this limitation, in this paper, we aim to equip anchors with more descriptive power to better approximate the input data $\bm{y}_i$ for making more accurate inferences from it.
We also provide an illustrative example in Fig.~\ref{manifold} for better understanding.
\section{The Proposed DDLCN Framework}
\label{formulation}

We sequentially introduce each layer of the proposed DDLCN in this section.
Note that we only illustrate details of two-layer DDLCN for simplicity.
Extension of the proposed DDLCN to multiple layers is straight forward.

\noindent \textbf{Feature Extraction Layer.} 
We first adopt a feature extractor $F$ to extract a set of $m$-dimensional local descriptors $\bm{Y} {=} \left[ {{\bm{y}_1}, \cdots ,{\bm{y}_l}} \right] {\in} {\R^{m \times l}}$ from the input image $I$, where $l$ is the total number of local descriptors.
To highlight the effectiveness of the proposed method, we only use a single feature extractor in our experiment, i.e., Scale-Invariant Feature Transform (SIFT)~\cite{lowe2004distinctive}.
The SIFT descriptor has been widely used in dictionary learning~\cite{shen2015multi,yang2017top,kim2017modality,bao2016dictionary,yan2019stat}. 
However, one can always use multiple feature extractor to further improve performance.
Specifically, for the input image $I$, we extract the SIFT feature $\bm{y_i}$ by using the feature extractor $F$,  this process can be formulated as $\bm{y}_i {=} F(I), i {\in} [1, ..., l]$.

\noindent \textbf{First Dictionary Learning Layer.} 
Let $r$ denote  the total number of classes in the dataset.
Then we randomly select $p$ images in each class to train the dictionary of the corresponding class, and the number of the first layer dictionary per category  is denoted as $q$.
Thus, the number of the dictionary for the first layer can be calculated by $D_1  {=} r {*} q$.
Next, we adopt the following dictionary learning algorithm,
\begin{equation}
\begin{array}{c}
\min\limits_{\bm{V}_i} \left[ \frac{1}{2} ||\bm{y}_i - \bm{V_i}\alpha_i||_2^2 \right]
\quad s.t.\quad ||\alpha_i||_1 <= \lambda
\end{array}
\end{equation}
where $\bm{V}_i {=} \left[ {{\bm{v}_1},{\bm{v}_2}, \cdots ,{\bm{v}_{{q}}}} \right]$ is the dictionary for $i^{th}$ class in the first-layer dictionary, which contains $q$ atoms, i.e.,~$\bm{v}_i$.
We then group all of them to form  the first-layer dictionary $\bm{V}$ after separately learning the dictionary of each class.
Thus $\bm{V} {=} \left[ {{\bm{V}_1},{\bm{V}_2}, \cdots ,{\bm{V}_{{r}}}} \right] {=} \left[ {{\bm{v}_1},{\bm{v}_2}, \cdots ,{\bm{v}_{{D_1}}}} \right] {\in} {\R^{m \times {D_1}}}$. $\alpha_i$ is a sparse coefficient introduced in \cite{mairal2009online}.
In this way, the dictionary $\bm{V}_i$ and the coefficients~$\alpha_i$ can be learned jointly.

\begin{figure}[!t] \small
	\centering
	\includegraphics[width=1\linewidth]{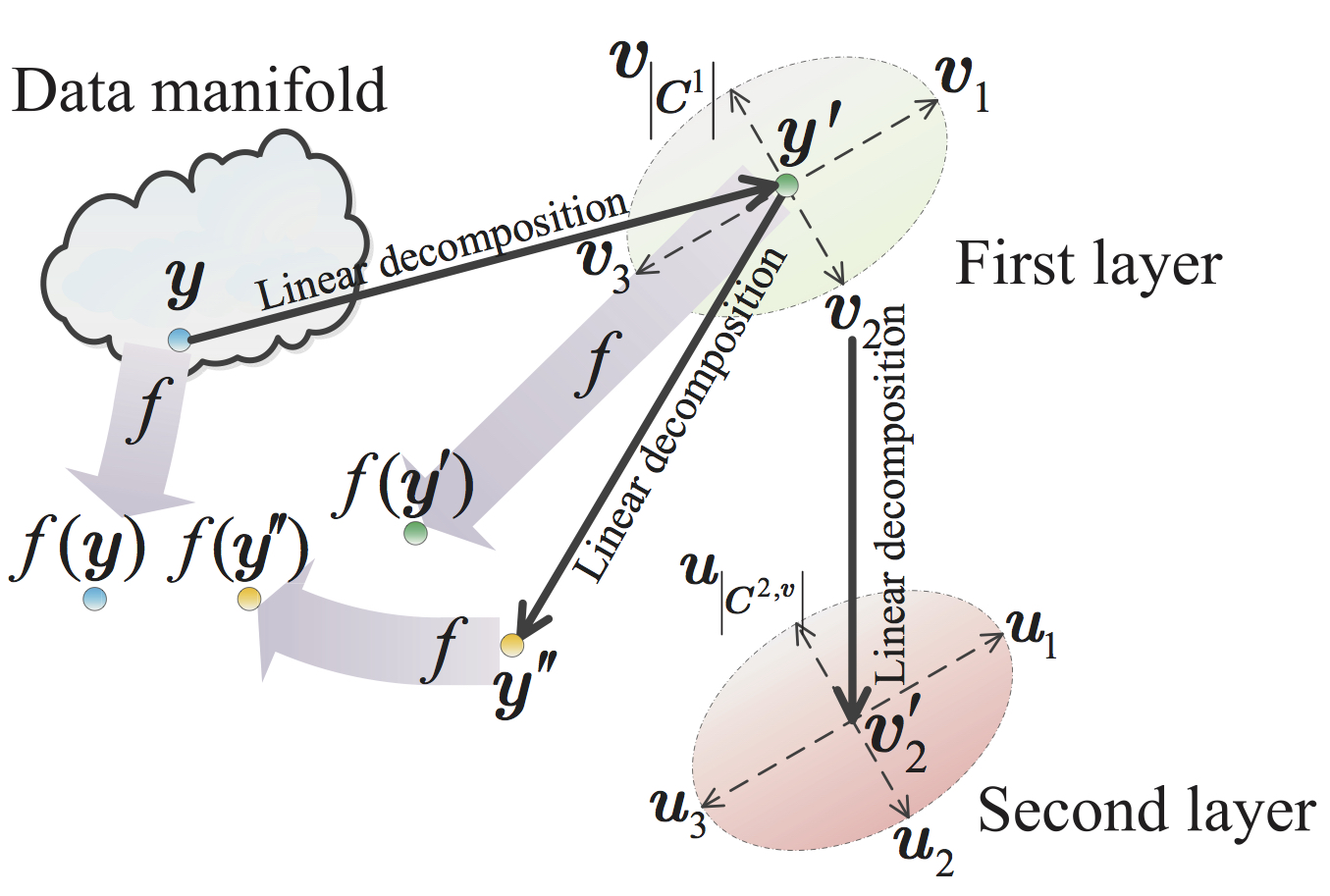}\\
	\caption{Multi layers coding strategy. The first layer is mainly used to partition the space, while the main approximation power is achieved within the second layer, which embodies a `divide and conquer' strategy.}
	\label{two_layer}
	\vspace{-0.4cm}
\end{figure}

\noindent \textbf{First Feature Coding Layer.}
After learning $\bm{V}$, each local feature is then encoded by $\bm{V}$ through several nearest atoms for generating the first coding.
By doing so, the first feature coding layer transfers each local descriptor $\bm{y}_i$ into a $D_1$ dimensional code ${\bm{\gamma} ^1} {=} \left[ {\bm{\gamma} _1^1,\bm{\gamma} _2^1, \cdots ,\bm{\gamma}_l^1} \right] {\in} {\R^{{D_1} \times l}}$.
Specifically, each code can be obtained using the following optimization,
\begin{equation}
\label{first_layer_formulation}
\begin{array}{c}
\mathop {\min }\limits_{\bm{\gamma} _i^1} \left[ \sum\limits_{i=1}^{l}{\frac{1}{2}\left\| {{\bm{y}_i} - \bm{V}\bm{\gamma} _i^1} \right\|_2^2 + \beta {{\left\| {\bm{\gamma} _i^1 \odot {\bm{\zeta}_i^1}} \right\|}_1}} \right] \\
\quad s.t.\quad {\bm{1}^\mathsf{T}}\bm{\gamma} _i^1 = 1,
\end{array}
\end{equation}
where $\bm{\zeta}_i^1 {\in} \R^{D_1}$ is a distance vector to measure the distance between $\bm{y}_i$ and $\bm{v}_i$. $ \odot $ denotes the element-wise multiplication.
Typically, $\bm{\zeta}_i^1$ can be obtained  by reducing a reconstruction loss in the corresponding layer.
We note that \cite{lin2010deep} adopts a simple sparse coding model at the first layer, which overlooks
the importance of quantity distributions of each item in the code $\bm{\gamma} _i^1$, thus it is prone to a rough approximation at the first layer. Therefore, the physical approximation of $\bm{y}$ in the first layer can be expressed as,
\begin{equation}
\label{h1}
\bm{y'} = \sum\limits_{\bm{v} \in {\bm{C}^1}} {\bm{\gamma}^1(\bm{y})\bm{v}},
\end{equation}
where ${\bm{C^1}}$ is the set of anchor points to $\bm{y}$. An illustrative example is shown in Fig.~\ref{two_layer}.

\noindent \textbf{Second Dictionary Learning Layer.} 
As discussed in the introduction, most existing dictionary learning frameworks only use a single layer, which significantly limits the discriminative ability of the feature coding. 
Meanwhile, we observe that better representation will be obtained by using deeper layers in most computer vision tasks. Thus, we borrow some idea from deep CNNs and present a new deeper dictionary learning and coding layer. Then the second layer dictionary $\bm{U} {=} \left[ {{\bm{u}_1},{\bm{u}_2}, \cdots ,{\bm{u}_{{s_2}}}} \right]$ can be learned from  the first layer dictionary $\bm{V}$,
\begin{equation}
\begin{array}{c}
\min\limits_{\bm{U}} \left[\frac{1}{2} ||\bm{v}_i - \bm{U}\alpha_i||_2^2 \right]
\quad s.t.\quad ||\alpha_i||_1 <= \lambda
\end{array}
\end{equation}
where ${\bm{v}_i}{\in}\bm{V}$ is one of the basis vectors in the first activated dictionary.
At the second layer, we put more emphasis on the representation of each $\bm{v}_i$ or each group of $\bm{v}_i$ to further refine each basis $\bm{v}_i$.
Specifically, after coding at the first layer, we try to map a nonlinear function $f$ to a simplified local coordinate space with low intrinsic dimensionality.
However, from the viewpoint of Lipschitz smoothness~\cite{yu2009nonlinear, lin2010deep}, this solo layer mapping only incorporates limited information about $f$ with its derivative on $\bm{y}$, such that it is incapable of guaranteeing better approximation quality.
That is why we would move deeper into the second layer to seek more information about $f$ for further improving the approximation.
By doing so, the first layer can capture the fine low-level structures from the input image, then the second coherently captures more complex structures from the first layer.

\noindent \textbf{Second Feature Coding Layer.}
We can obtain the code of  the second layer by using the following optimization, 
\begin{equation}
\label{formulation_second_layer}
\begin{array}{c}
\min\limits_{\bm{\gamma} _i^{2}} \left[ \sum\limits_{i=1}^{D_1} {\frac{1}{2}\left\| {{\bm{v}_i} - \bm{U}\bm{\gamma} _i^{2}} \right\|_2^2 + \beta {{\left\| {\bm{\gamma} _i^{2} \odot \bm{\zeta}_i^2} \right\|}_1}} \right] \\
\quad s.t.\quad {1^\mathsf{T}}\bm{\gamma} _i^{2} = 1,
\end{array}
\end{equation}
where $\bm{\gamma}_i^{2} {=} {\left[ {\bm{\gamma} _i^{2}({\bm{u}_1}),\bm{\gamma} _i^{2}({\bm{u}_2}), \cdots ,\bm{\gamma} _i^{2}({\bm{u}_{{D_2}}})} \right]^\mathsf{T}} \linebreak[3] {\in} {\R^{{D_2}}}$ is the second coding and $D_2$ is the number dictionary of the second layer. 
$\bm{\zeta}_i^2 {\in} {\R^{{D_2}}}$ is used to measure the distance between $\bm{v}_i$ and each atom in $\bm{U}$.
$\bm{v}_i {\in} \bm{V}$ is one of the basis vectors adopted in the representation of $\bm{y}_i$ at the first layer.

By doing so, the activated atoms $\bm{v}_i$ in the first layer can be further decomposed to obtain the second layer coding using $\bm{U}$.
Thus, the approximation of $\bm{y}$ in the second layer can be defined as,
\begin{equation}
\label{h2}
\bm{y}'' = \sum\limits_{\bm{v} \in {\bm{C}^1}} {\left[ {\bm{\gamma}^1(\bm{y})\sum\limits_{\bm{u} \in {\bm{C}^{2,\bm{v}}}} {\bm{\gamma}^{2,\bm{v}}(\bm{v})\bm{u}} } \right]},
\end{equation}
where ${\bm{C^{2,v}}}$ is the set of anchor points to $\bm{v}$. 
We also provide an illustrative example in Fig.~\ref{two_layer} for better understanding.
The core idea of the two-layer coordinate coding is that if both coordinate codings, i.e., $\bm{y}'$ and $\bm{v}' {=} \sum\limits_{\bm{u} \in {\bm{C}^{2,\bm{v}}}} {\bm{\gamma} ^{2,\bm{v}}(\bm{v})\bm{u}}$, are sufficiently localized, then a point $\bm{y}$ lies on a manifold, which would be locally embedded into a lower-dimensional two-layer structure space.
More importantly, not only the data point $\bm{y}$ is locally linearly represented, but also the function $f(\bm{y})$.
This significant observation lays the foundation for our approach.

\noindent \textbf{The $n^{th}$ Dictionary Learning Layer.} 
Similarly, we can learn the $n^{th}$ dictionary $\bm{D^{n}} {=} \left[ {{\bm{d}_1^{n}},{\bm{d}_2^{n}}, \cdots ,{\bm{d}_{{D_n}}^{n}}} \right]$ from the previous layer dictionary $\bm{D^{n-1}}$,
\begin{equation}
	\begin{array}{c}
		\min\limits_{\bm{D^{n}}} \left[ \frac{1}{2} ||\bm{d}_i^{n-1} - \bm{D^{n}}\alpha_i||_2^2 \right]
		\quad s.t.\quad ||\alpha_i||_1 <= \lambda,
	\end{array}
\end{equation}
where ${\bm{d}_i^{n-1}}{\in} \bm{D^{n-1}}$ is one of the activated basis vectors in the previous $(n{-}1)^{th}$ dictionary layer.   

\noindent \textbf{The $n^{th}$ Feature Coding Layer.}
Therefore, we can generalize the two-layer framework of DDLCN to a deeper one,
\begin{equation}
\begin{array}{c}
\min\limits_{\bm{\gamma} _i^{n}} \left[ {\frac{1}{2}\left\| {\bm{d}_i^{n-1} - \bm{D^n}\bm{\gamma}_i^{n}} \right\|_2^2 + \beta {{\left\| {\bm{\gamma} _i^{n} \odot \bm{\zeta}_i^n} \right\|}_1}} \right] \\
\quad s.t.\quad {1^\mathsf{T}}\bm{\gamma}_i^{n} = 1,
\end{array}
\end{equation}
where $\bm{\gamma} _i^{n}$ is the $n^{th}$ layer coding and 
$\bm{\zeta}_i^n$ is employed to measure the distance between $\bm{d}_i^{n-1}$ and each atom in $\bm{D^n}$.
$\bm{d}_i^{n-1} {\in} \bm{D^{n-1}}$ is one of the basis vectors adopted in the feature representation of $\bm{y}_i$ at the $(n{-}1)^{th}$ coding layer.
Through the proposed multi-layer learning and coding strategy, the proposed DDLCN can output a robust feature representation to accurately represent the input image.
Moreover, DDLCN increases and boosts the separability of feature representations from different semantic classes.
Lastly, DDLCN preserves the locality information of the input local features, avoiding very large values in the coding representation and  reducing the error caused by over-fitting.

\noindent \textbf{Pooling Layer.}
After the last dictionary learning and feature coding layer, we use a pooling layer for removing the fixed-size constraint of the input images \cite{he2015spatial}.
Specifically, for each input image, we adopt $1{\times}1$, $2{\times}2$ and $4{\times}4$  spatial pyramids with max-pooling.

\noindent \textbf{Fully Connected Layer.}
The final feature representations of  $\bm{y}_i$ can be obtained by integrating feature representation from each layer. 
Task two-layer framework for an example, 
each item (such as the $j^{th}$ item) in the first layer's codes $\bm{\gamma} _i^1$ can be augmented  into the form of ${\left[ {\bm{\gamma}_i^1({\bm{v}_j}), \bm{\gamma} _i^1({\bm{v}_j})[\bm{\gamma} _j^{2}({\bm{u}_1}),\bm{\gamma} _j^{2}({\bm{u}_2}), \cdots ,\bm{\gamma} _j^{2}({\bm{u}_{{s_2}}})]} \right]^\mathsf{T}}$. 
Then we concatenate the first layer coding and the second layer coding to form the final coding representation, which is a $D_1 {\times} (1 {+} D_2)$ dimensional vector.
We also provide the two-layer framework of our DDLCN in Algorithm~\ref{alg:SA}.

\noindent \textbf{Output Layer.}
We adopt the Support Vector Machine (SVM) \cite{cortes1995support} as our classifier, which has been validated in many classification tasks such as \cite{wang2010locality,tang2015gender,song2019multi,tang2019fast,van2008kernel}. 
Specifically, we employ LIBSVM~\cite{chang2011libsvm} to  implement our multi-class SVM.

The classification of the input image is ultimately carried out by assembling deep dictionaries from different layers and assessing their contribution.
Moreover, through jointly minimizing both the classification errors and the reconstruction errors of all different layers, the proposed DDLCN iteratively adapts the deep dictionaries to help to build better feature representations for image recognition tasks.

\begin{algorithm}[!t] \small
	\caption{The two-layer framework of our DDLCN.}
	\label{alg:SA}
	\begin{algorithmic}[1]
		\REQUIRE:
		$\bm{Y} \in {\R^{m \times l}}$
		\ENSURE:
		$\bm{\gamma} _i$\\
		\STATE First dictionary learning: $ \quad\bm{V}\leftarrow\bm{V}_{Dictionary}$
		\STATE First locality constraint calculating:\\
		${\quad\bm{\zeta}_i^1} = \left[ {{{\left\| {{\bm{y}_i} - {\bm{v}_1}} \right\|}_2},{{\left\| {{\bm{y}_i} - {\bm{v}_2}} \right\|}_2}, \cdots ,{{\left\| {{\bm{y}_i} - {\bm{v}_{{D_1}}}} \right\|}_2}} \right]^\mathsf{T}$
		\STATE First feature coding:\\
		$\quad \textbf{for}\quad i=1$ to $l$\\
		$\quad\begin{array}{l}
		\bm{\gamma} _i^1 \leftarrow \mathop {\min }\limits_{\bm{\gamma} _i^1} \left[ {\frac{1}{2}\left\| {{\bm{y}_i} - \bm{V}\bm{\gamma} _i^1} \right\|_2^2 + \beta {{\left\| {\bm{\gamma} _i^1 \odot {\bm{\zeta}_i^1}} \right\|}_1}} \right] \\
		\quad \quad s.t.\quad {1^\mathsf{T}}\bm{\gamma} _i^1 = 1
		\end{array}$\\
		$\quad \textbf{end}$
		
		\STATE Second dictionary learning: $\quad\bm{U}\leftarrow\bm{U}_{Dictionary}$
		\STATE Second locality constraint calculating:\\
		$\quad{\bm{\zeta}_i^2} = \left[ {{{\left\| {{\bm{v}_i} - {\bm{u}_1}} \right\|}_2},{{\left\| {{\bm{v}_i} - {\bm{u}_2}} \right\|}_2}, \cdots ,{{\left\| {{\bm{v}_i} - {\bm{u}_{{D_2}}}} \right\|}_2}} \right]^\mathsf{T}$
		\STATE Second feature coding:\\
		$\quad \textbf{for}\quad i=1$ to $D_1$\\
		$\quad \begin{array}{l}
		\bm{\gamma} _i^{2} \leftarrow \mathop {\min }\limits_{\bm{\gamma} _i^{2}} \left[ {\frac{1}{2}\left\| {{\bm{v}_i} - \bm{U}\bm{\gamma} _i^{2}} \right\|_2^2 + \beta {{\left\| {\bm{\gamma} _i^{2} \odot \bm{\zeta}_i^2} \right\|}_1}} \right] \\
		\quad \quad s.t.\quad {1^\mathsf{T}}\bm{\gamma} _i^{2} = 1
		\end{array}$\\
		$\quad \textbf{end}$	
		\STATE Coding augmentation:\\
		$\quad \textbf{for}\quad i=1$ to $l$\\
		$\quad\quad \textbf{for}\quad j=1$ to $D_1$ \\
		$\quad\quad\quad {{\bm{\gamma}_i^1({\bm{v}_j})} \leftarrow\left[ {\bm{\gamma} _i^1({\bm{v}_j}), \bm{\gamma}_i^1({\bm{v}_j})[\bm{\gamma} _j^{2}({\bm{u}_1}),\cdots ,\bm{\gamma} _j^{2}({\bm{u}_{{D_2}}})]} \right]^\mathsf{T}}$ \\
		$\quad\quad \textbf{end}$ \\
		$\quad \textbf{end}$
	\end{algorithmic}
\end{algorithm}
\section{Experiments}
\label{experiment}
We conduct extensive experiments (including face recognition, object recognition and hand-written digits recognition) to evaluate the effectiveness of the proposed DDLCN.

\subsection{Experimental Setting}
\noindent \textbf{Datasets.}
We follow~\cite{bao2016dictionary,akhtarjoint,tang2018deep,wu2019structured,zhang2017jointly,hu2018nonlinear,xiao2015two,yan2019cross} and  evaluate the effectiveness of the proposed DDLCN on five widely-used datasets, i.e., Extended YaleB~\cite{georghiades2001few}, AR Face~\cite{MARTINEZ.M.:1998}, Caltech 101~\cite{fei2007learning}, Caltech 256~\cite{Griffin07tech} and MNIST~\cite{lecun1998gradient}), which are all standard datasets for dictionary learning evaluation. 
Note that we follow the same evaluation procedure with the previous works on each dataset for a fair comparison. 

\begin{figure}[!t] \small
	\begin{centering}
		\centering
		\setcounter{subfigure}{0}
		\subfigure{\includegraphics[width=0.49\linewidth]{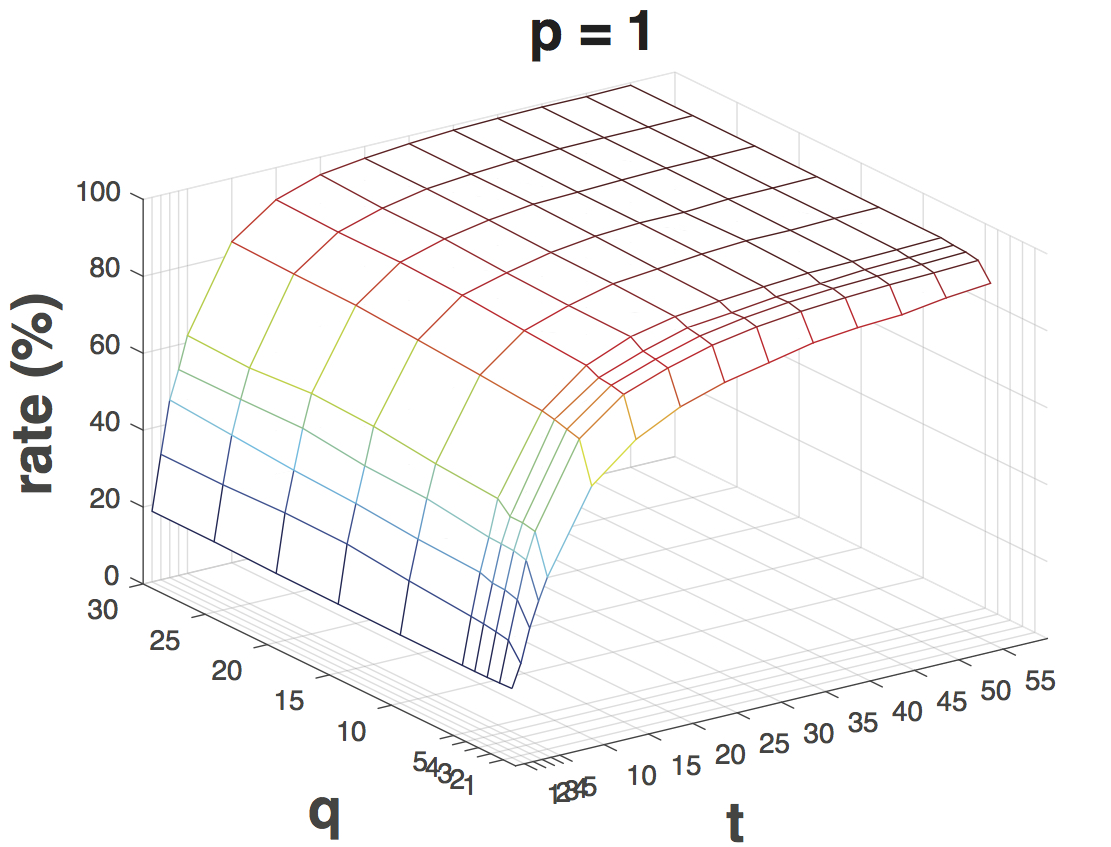}}
		\subfigure{\includegraphics[width=0.49\linewidth]{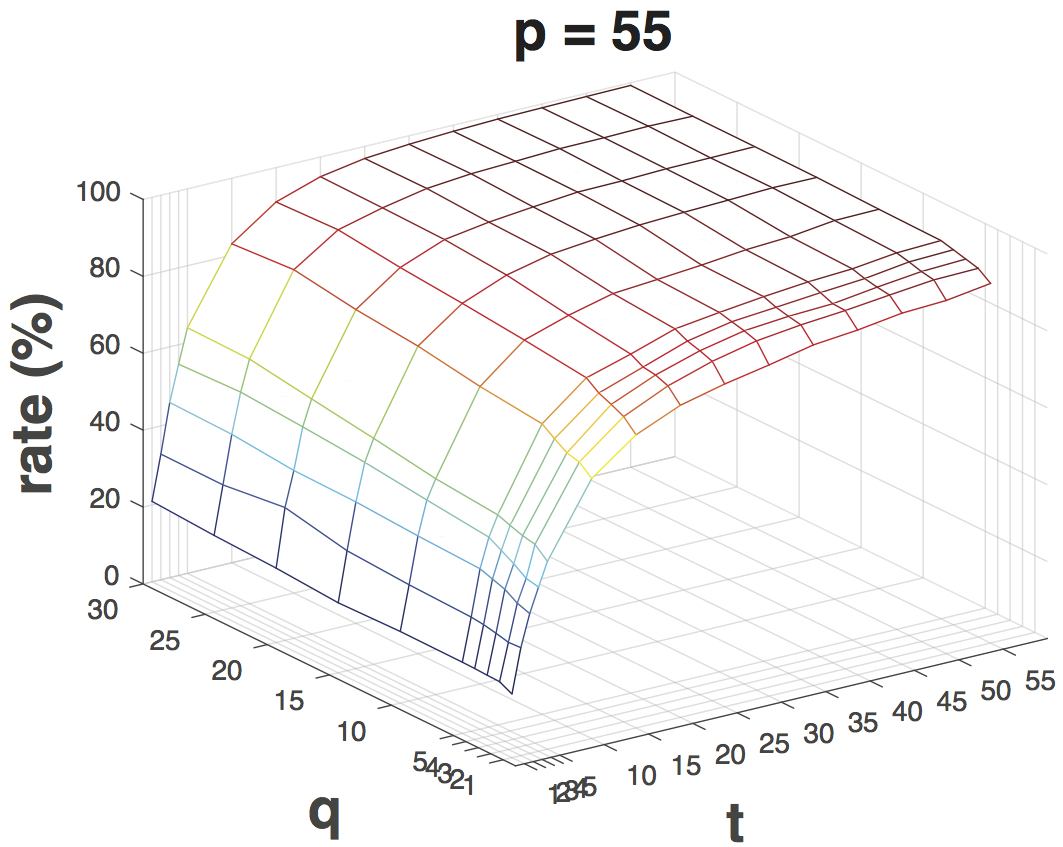}}
		
		\subfigure{\includegraphics[width=0.49\linewidth]{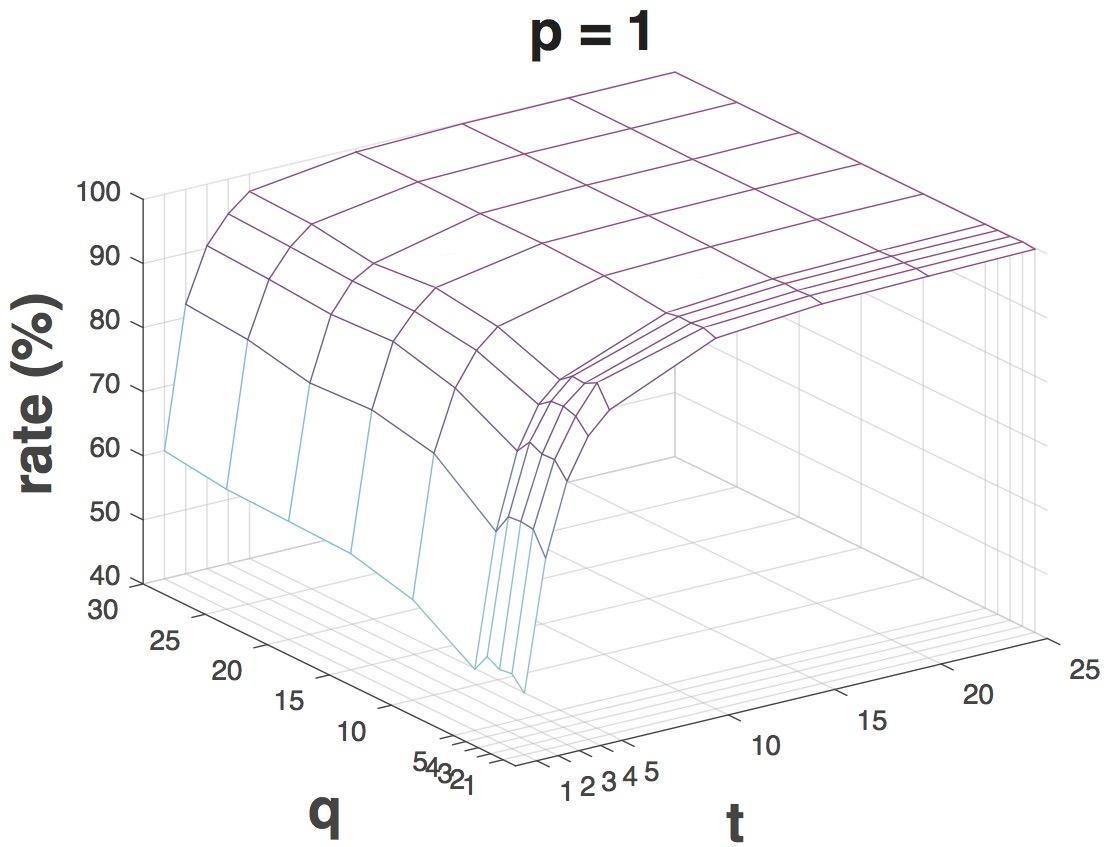}}
		\subfigure{\includegraphics[width=0.49\linewidth]{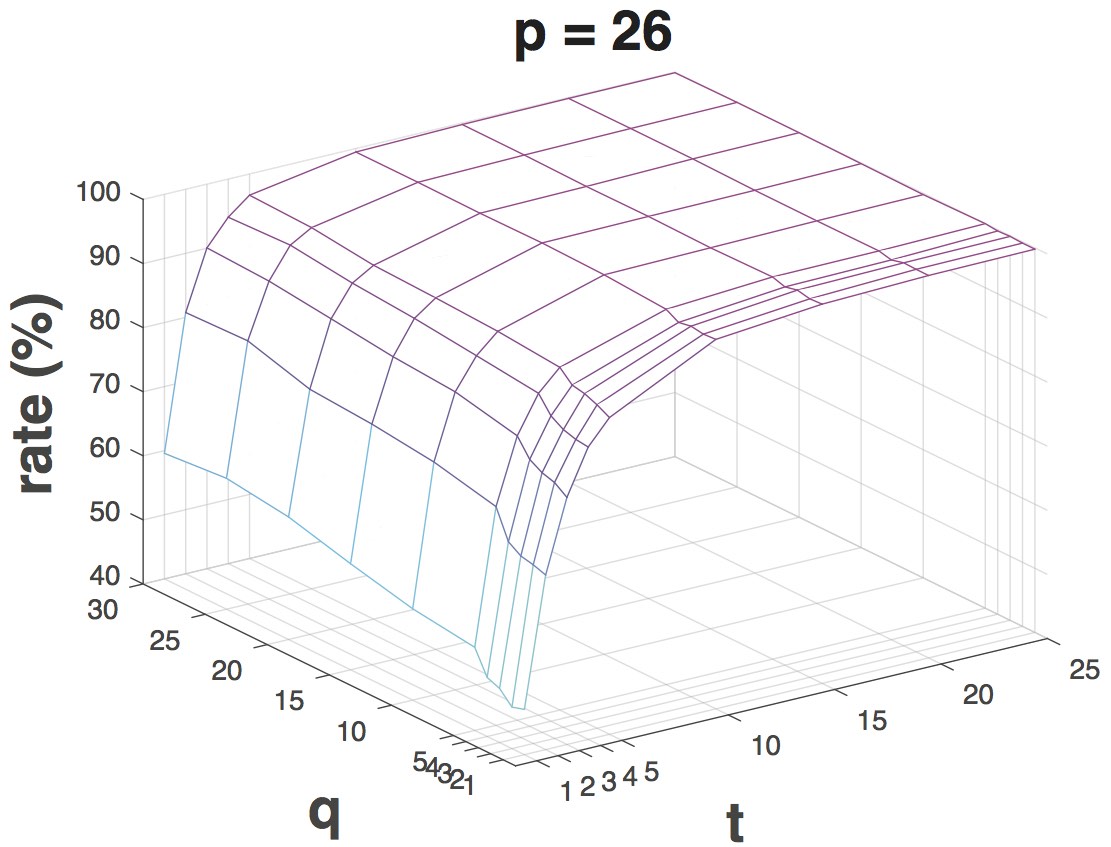}}
		\caption{Classification accuracy with different parameter $p$ on Extended YaleB (\textit{top two}) and AR Face (\textit{bottom two}).}
		\label{Fig:yaleb_ar_triandictsamp}
	\end{centering}
	\vspace{-0.4cm}
\end{figure}

\begin{figure}[!t] \small
	\begin{centering}
		\centering
		\setcounter{subfigure}{0}
		\subfigure{\includegraphics[width=0.49\linewidth]{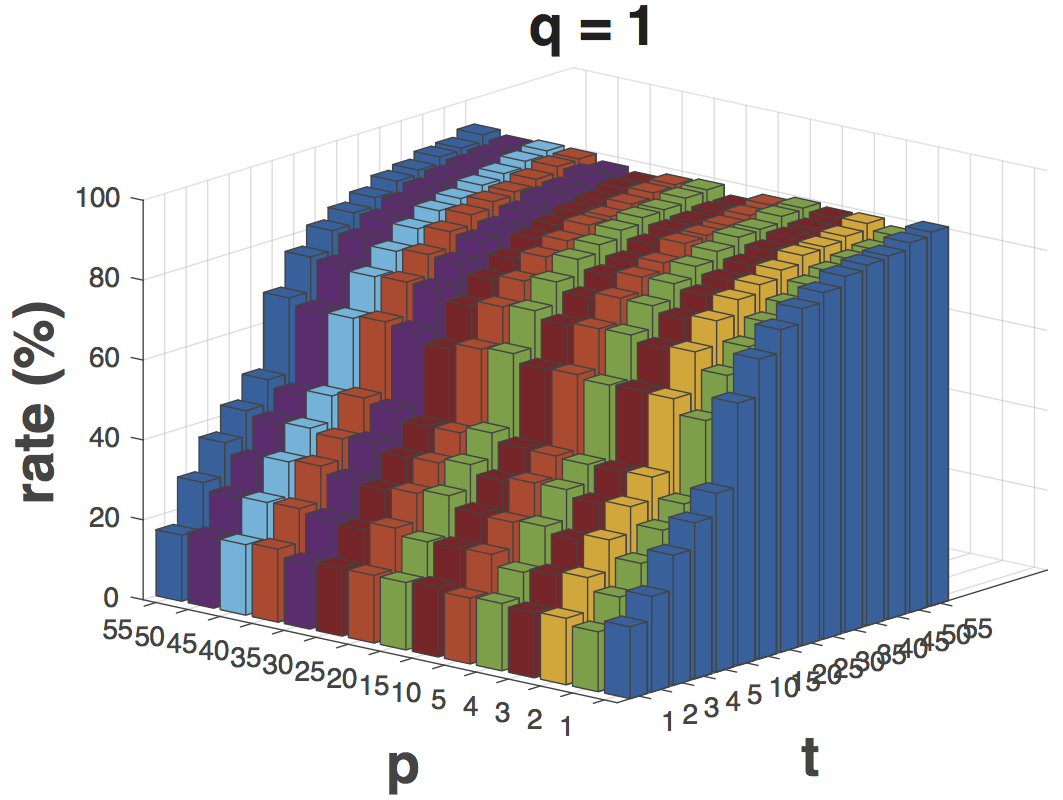}}
		\subfigure{\includegraphics[width=0.49\linewidth]{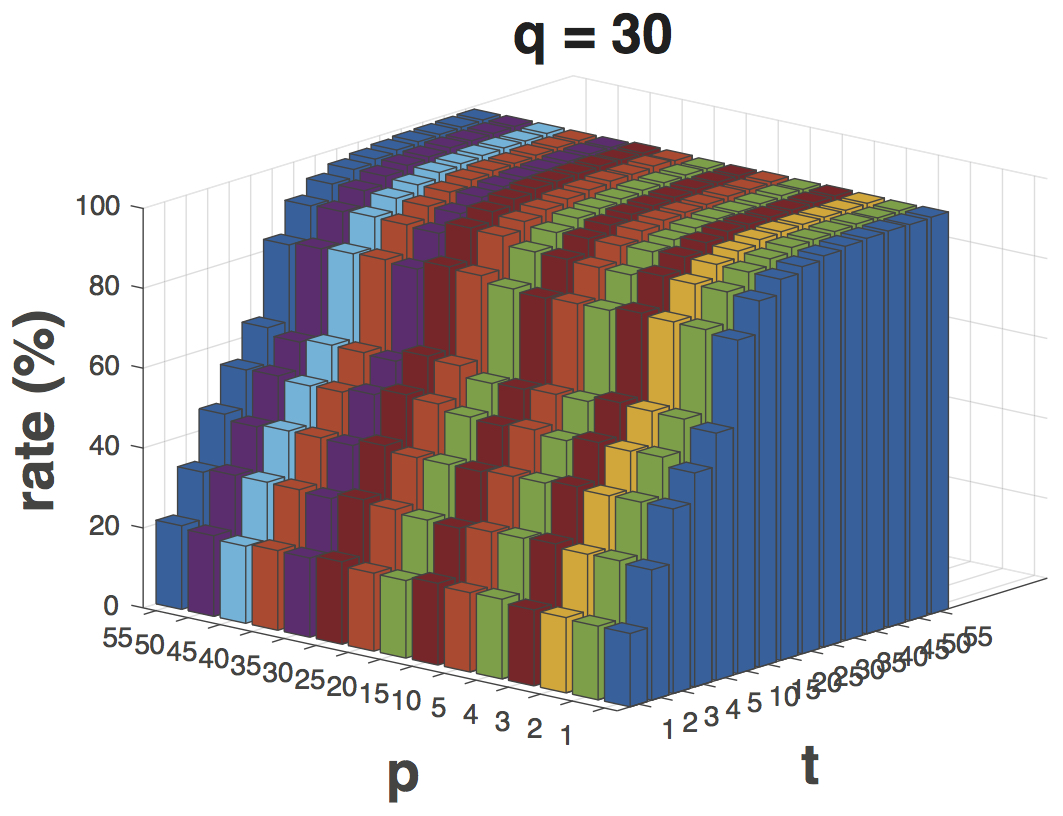}}
		
		\subfigure{\includegraphics[width=0.49\linewidth]{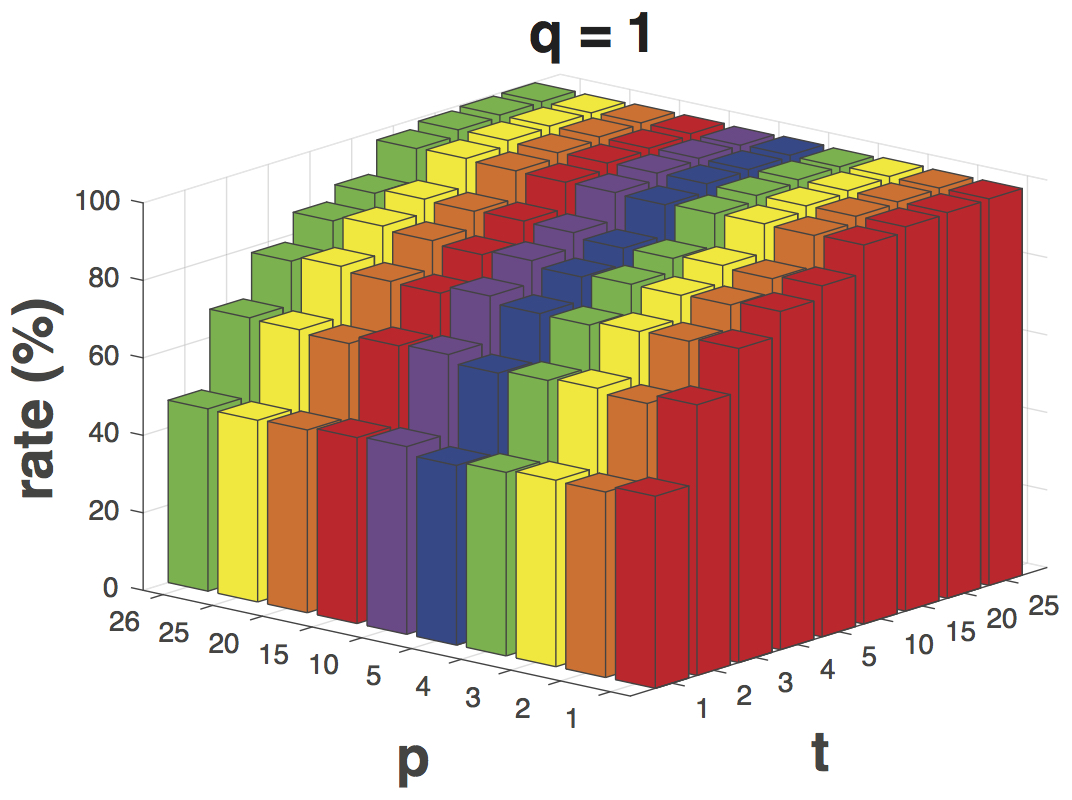}}
		\subfigure{\includegraphics[width=0.49\linewidth]{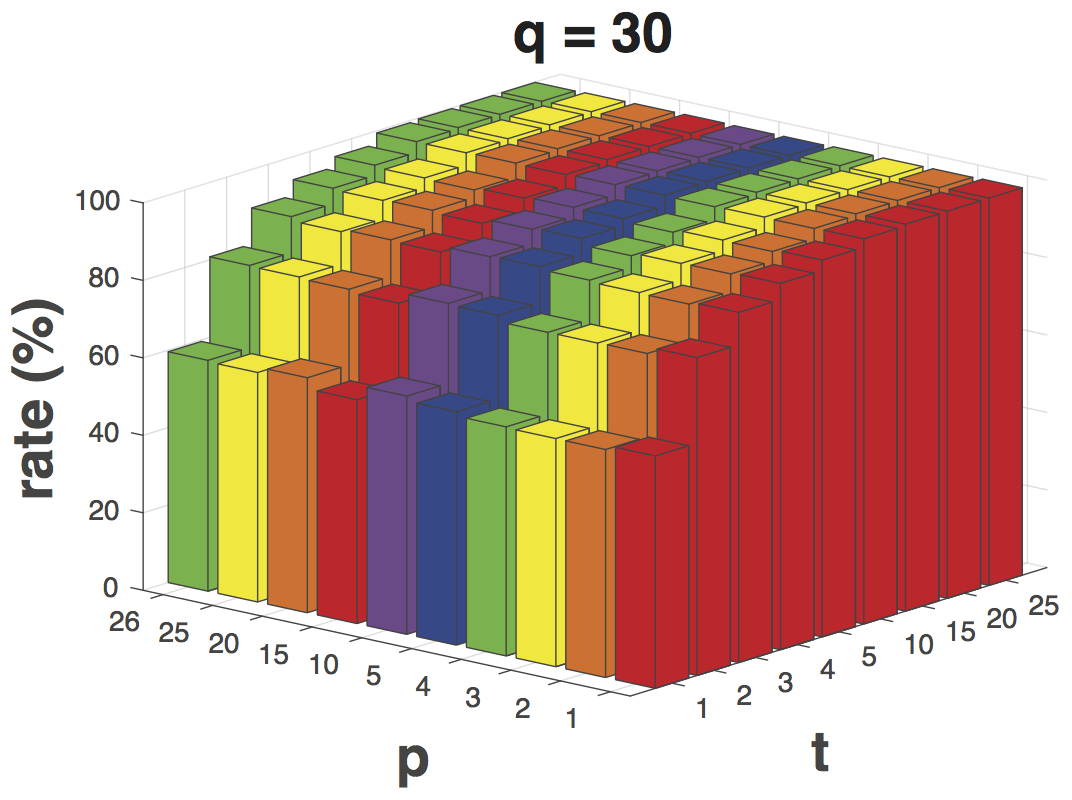}}	
		\caption{Classification accuracy with varying $q$ on Extended YaleB (\textit{top two}) and AR Face (\textit{bottom two}).}
		\label{Fig:yaleb_ar_firstdictsize}
	\end{centering}
	\vspace{-0.4cm}
\end{figure}

\noindent \textbf{Parameter Setting.}
Compared with existing CNN-based methods, which have lots of hyper-parameters, while the proposed DDLCN only has three parameters needed to be tuned. 
The three parameters of the proposed DDLCN are:
\begin{itemize}
	\item $p$, the number of training dictionary samples per category.
	\item $q$, the number of the first layer dictionary per category.
	\item $t$, the number of training samples per category.
\end{itemize}
For simplicity, we use $p{-}q$ to represents that $p$ images are randomly selected per category for training dictionary and $q$ dictionary bases are learned per category in the first dictionary. 
For example, `5-5' means that $p{=}5$ and $q{=}5$.
We first conduct extensive experiments to evaluate the performance of different values of $p$, $q$ and $t$.
For all experiments, we repeat 10 times to achieve reliable results and then average them to obtain the final results.

(i) Parameter $p$.
To demonstrate  the superiority of the proposed DDLCN, we set the number of the first dictionary to $p{=}[1,2,3,4,5,10,15,20,25,30,35,40,45,50,55]$ and $p{=}[1,2,3,4,5,10,15,20,25,26]$ on Extended YaleB and AR Face, respectively.
The results of two extreme cases on both datasets are shown in Fig.~\ref{Fig:yaleb_ar_triandictsamp}.
We can see that the proposed method achieves good results when $p{=}1$, validating our design motivation.
Moreover, we observe that the classification performance achieves a peak with 10 training samples and then tends to be stable.

(ii) Parameter $q$.
We set $q {=} [1,2,3,4,5,10,15,20,25,30]$ on both Extended YaleB and AR Face datasets.
The results of two extreme cases on both datasets are reported in Fig~\ref{Fig:yaleb_ar_firstdictsize}.
We can see that the gaps between the two cases are marginal due to the introduction of the proposed layer on Extended YaleB.
Moreover, the proposed DDLCN achieves nearly 100\% classification accuracy when only using 1 atom per person and 20 images per class for training on AR Face. 
This means that the proposed DDLCN can exploit the intrinsic structure of the manifold where features reside, leading to better classification performance with limited dictionaries and training data.

\begin{figure}[!t] \small
	\centering
		\setcounter{subfigure}{0}
		\subfigure{\includegraphics[width=0.49\linewidth]{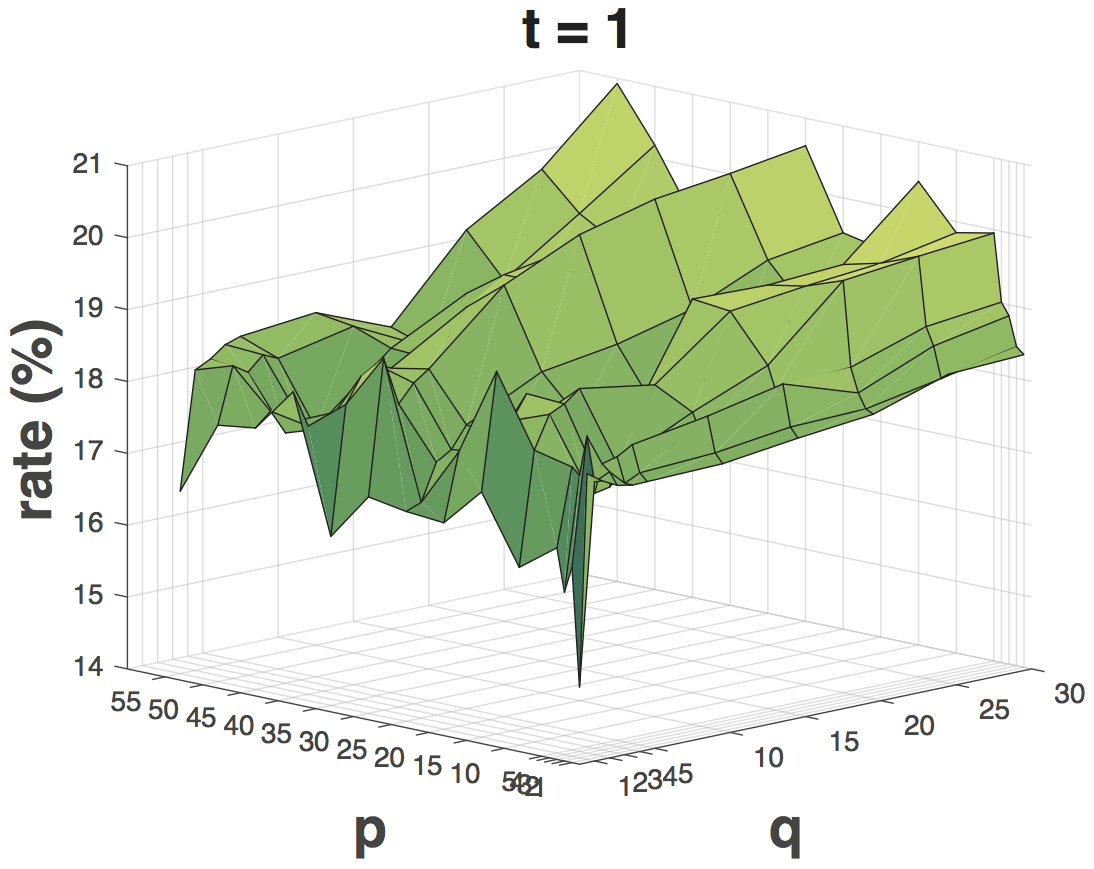}}
		\subfigure{\includegraphics[width=0.49\linewidth]{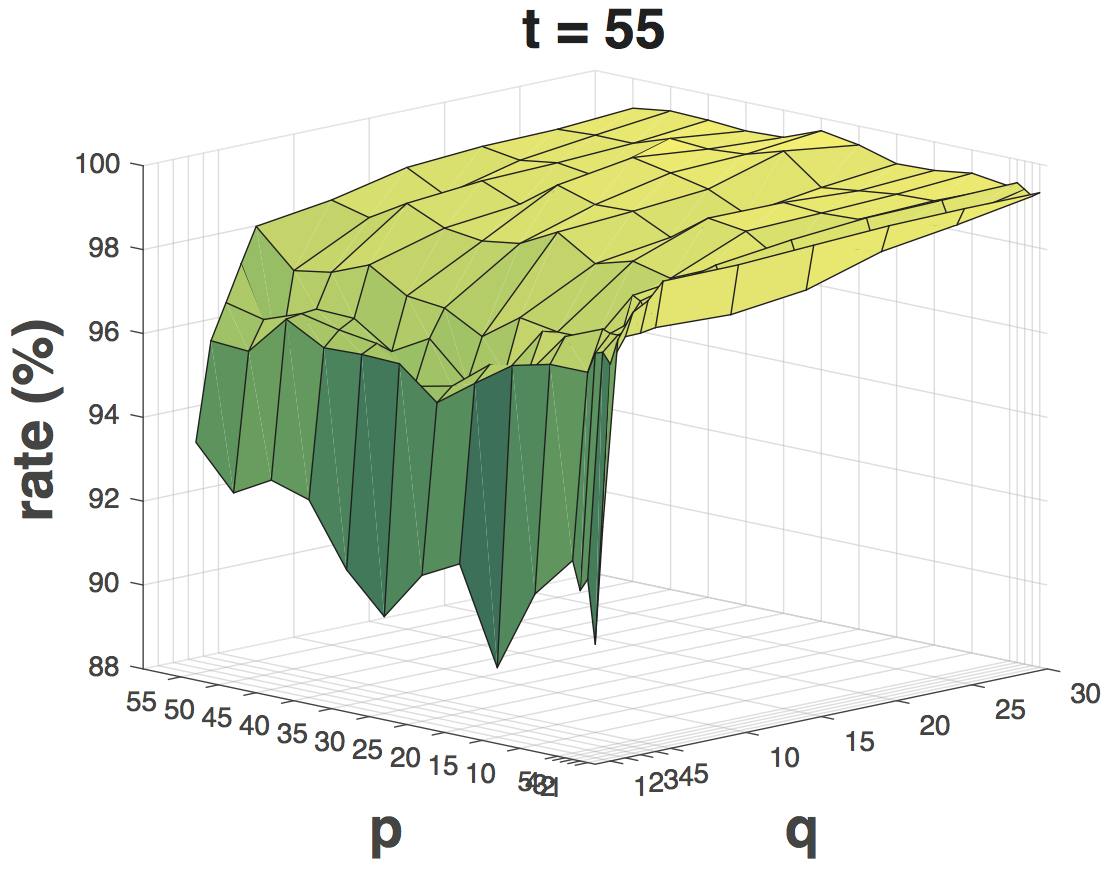}}
		
		\subfigure{\includegraphics[width=0.49\linewidth]{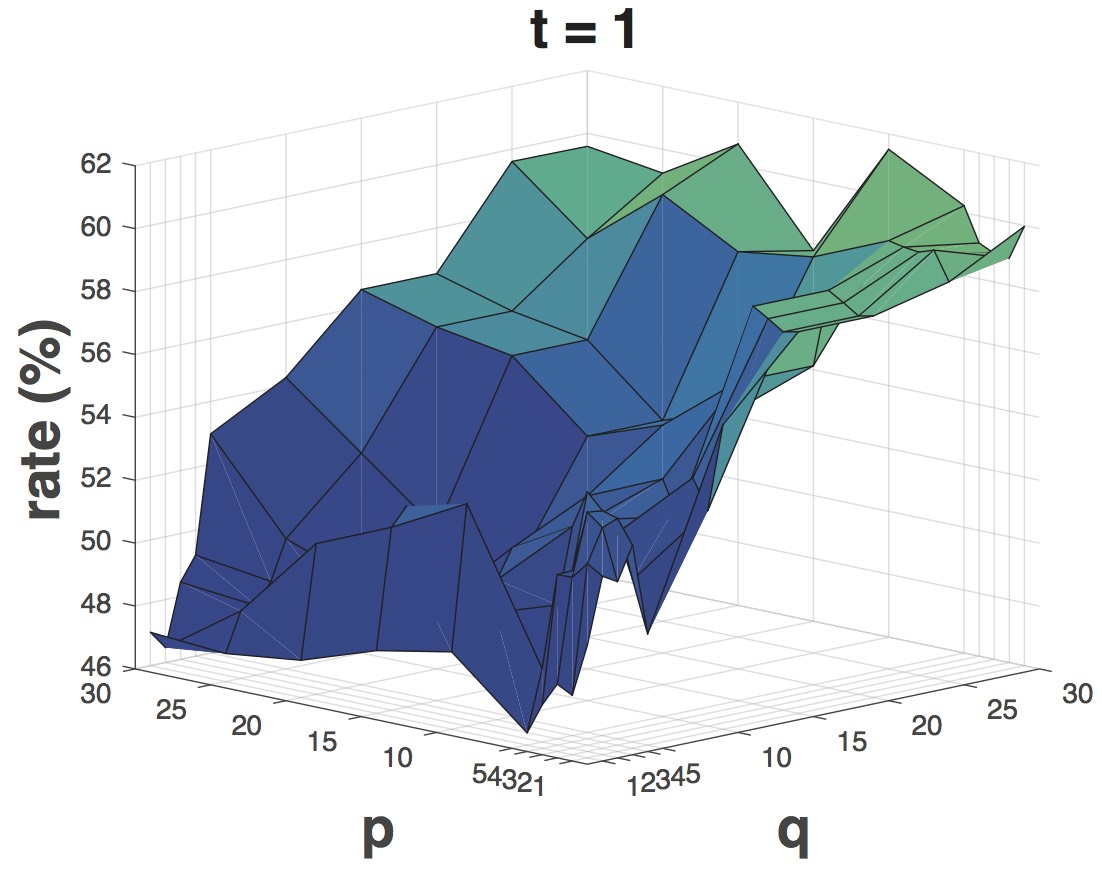}}
		\subfigure{\includegraphics[width=0.49\linewidth]{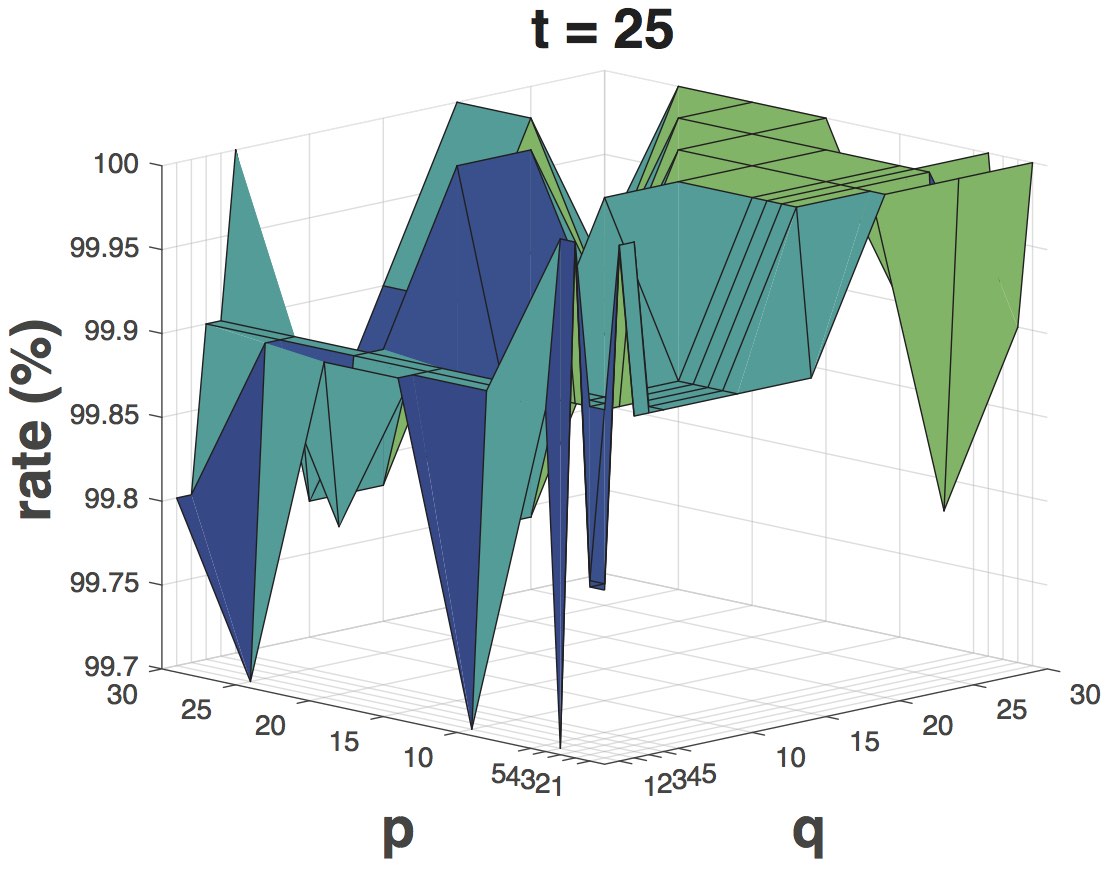}}
		\caption{Classification accuracy with different $t$ on Extended YaleB (\textit{top two}) and AR Face (\textit{bottom two}).}
		\label{Fig:yaleb_ar_trainsamp}
			\vspace{-0.4cm}
\end{figure}

(iii) Parameter $t$.
We set the number of the training images to $t {=} [1,2,3,4,5,10,15,20,25,30,35,40,45,50,55]$ and $t {=} [1,2,3,4,5,10,15,20,25]$ on Extended YaleB and AR Face, respectively. 
The results of two extreme cases on both datasets are illustrated in Fig.~\ref{Fig:yaleb_ar_trainsamp}.
We can draw two conclusions:
1) the classification accuracy first rises to the peak rapidly and then tends to be stable as $t$ increasing.
2) there is a small impact to classification accuracy when changing $p$.

\begin{table}[!t] \small
	\centering
	\caption{Comparison results between the proposed  dictionary layer and conventional convolutional layer.}
	\resizebox{\linewidth}{!}{
		\begin{tabular}[!hb]{l|c|c}
			\hline  Method                                & Extended YaleB    & AR Face    \\ \hline
			DDLCN-2 (Dictionary  Layer)  & \textbf{99.18 $\pm$ 0.46}  & \textbf{99.87 $\pm$ 0.19}             \\ 
			DDLCN-2 (Convolutional Layer)                  & 98.94 $\pm$ 0.57  & 99.59 $\pm$ 0.38 \\  \hline
			DDLCN-3 (Dictionary  Layer)  & \textbf{99.32 $\pm$ 0.41}  & \textbf{99.92 $\pm$ 0.15}            \\ 
			DDLCN-3 (Convolutional Layer)                 & 99.07 $\pm$ 0.53  & 99.67 $\pm$ 0.36 \\ 
			\hline
		\end{tabular}}
		\label{table:cov_comp}
		\vspace{-0.4cm}
\end{table}

\subsection{The Proposed Layer vs. Convolutional Layer}
We then conduct experiments to validate the effectiveness of the proposed compound dictionary learning and coding layer.
Specifically, we employ the proposed DDLCN as our backbone and replace the proposed compound layers with conventional convolutional layers keeping all other details the same. 
Comparison results of both Extended YaleB and AR Face datasets are shown in Table \ref{table:cov_comp}.
We can see that the proposed compound dictionary learning and coding layer achieves better results than the convolutional layer, meaning that the proposed layer indeed obtains a more informative and discriminative representation, and confirming our design motivation.

\begin{table}[!t] \small
	\centering
	\caption{Classification accuracy (\%) on Extended YaleB.}
	\resizebox{\linewidth}{!}{
		\begin{tabular}[!hb]{l|c|c|c}
			\hline  Method                                          & Included (\%)     & Excluded$^{\ast}$ (\%)     & Time (ms) \\ \hline
			SRC (15 per person) \cite{wright2009robust}             & 80.50             & 86.70                      & 11.22 \\
			LLC (30 local bases) \cite{wang2010locality}            & 82.20             & 92.10                      & -    \\
			DL-COPAR \cite{wang2014classification}                  & 86.47 $\pm$ 0.69  & -                          & 31.11 \\
			FDDL \cite{yang2014sparse}                              & 90.01 $\pm$ 0.69  & -                          & 42.48 \\
			LLC (70 local bases) \cite{wang2010locality}            & 90.70             & 96.70                      & -    \\
			DBDL \cite{akhtar2016discriminative}                    & 91.09 $\pm$ 0.59  & -                          & 1.07 \\
			JBDC \cite{akhtarjoint}                                 & 92.14 $\pm$ 0.52  & -                          & 1.02 \\
			K-SVD (15 per person) \cite{aharon2006k}                & 93.10             & 98.00                      & - \\
			SupGraphDL-L \cite{yankelevsky2017structure}            & 93.44             & -                          & - \\	
			D-KSVD (15 per person) \cite{zhang2010discriminative}   & 94.10             & 98.00                      & - \\
			LC-KSVD1 (15-15) \cite{jiang2011learning}               & 94.50             & 98.30                      & 0.52 \\
			LC-KSVD2 (15-15) \cite{jiang2011learning}               & 95.00             & 98.80                      & 0.49 \\
			MBAP~\cite{bao2016dictionary}& 95.12             & -                          & - \\ 
			VAE+GAN \cite{mathieu2016disentangling}               & 96.4              & -                          & - \\
			EasyDL \cite{quan2016sparse}                            & 96.22             & -                          & - \\	
			LC-KSVD2 (A-15) \cite{jiang2011learning}                & 96.70             & 99.00                      & - \\	
			SRC (all training samples) \cite{wright2009robust}      & 97.20             & 99.00                      & 20.78 \\
			MDDL-2~\cite{song2019multi}                                  & 98.2                      & -                         & - \\	
			MDDL-3~\cite{song2019multi}                                  & 98.3                      & -                         & - \\	
			DDL~\cite{mahdizadehaghdam2019deep}                 & 99.1                       & -                         & - \\
			RRC\_L$_1$ (300) \cite{yang2013regularized}             & 99.80             & -                          & - \\ \hline
			PCANet-1 \cite{chan2015pcanet}                          & 97.77             & -                          & - \\
			PCANet-2 \cite{chan2015pcanet}                          & \textbf{99.85}    & -                          & - \\ \hline
			DDLCN-2 (1-1)                                             & 87.42 $\pm$ 1.33  & 89.54 $\pm$ 1.02           & \textbf{0.18} \\
			DDLCN-2 (15-15)                                           & 97.38 $\pm$ 0.54  & 98.48 $\pm$ 0.48           & 0.71 \\
			DDLCN-2 (55-15)                                           & 97.68 $\pm$ 0.60  & 98.64 $\pm$ 0.52           & 0.92 \\
			DDLCN-2 (A-15)                                            & 98.34 $\pm$ 0.56  & 99.18 $\pm$ 0.46            & 0.98 \\ 
			DDLCN-3 (A-15)                                            & 98.76 $\pm$ 0.42  & \textbf{99.32 $\pm$ 0.41}  & 1.32 \\ \hline
	\end{tabular}}
	\label{table:extendedYaleB}
	\vspace{-0.4cm}
\end{table}

\subsection{Comparison Against State-of-the-Art Methods}
We compare the proposed DDLCN (i.e., DDLCN-2 and DDLCN-3) with both dictionary learning and deep learning methods on five public datasets, i.e., Extended YaleB, AR Face, Caltech 101, Caltech 256 and MNIST datasets.

Specifically, DDLCN-2 uses two compound dictionary learning and coding layers, the first layer aims to learn a dictionary to represent the input image and the second layer target to learn a dictionary to represent the activated atoms in the first layer.
For a given input image, we use both the first and second dictionaries to learn the coding representation and then concatenate both of them to obtain the final coding representation.	
For a deeper version, i.e., DDLCN-3, which uses three compound dictionary learning and coding layers.
The first layer aims to learn a dictionary to represent input image, then the second layer targets to learn a dictionary to represent the activated atoms of the first layer, finally, the third layer learns another dictionary to represent the activated atoms of the second layer. 
We use the first, second and third dictionaries to learn the corresponding coding representation and then concatenate all three to form the final coding representation. 
The only difference between DDLCN-2 and DDLCN-3 consists in the number of the proposed layer, while all other layers such as fully connected and pooling layers, and the training details are all the same.

\begin{table}[!t] \small
	\centering
	\caption{Classification accuracy (\%) on AR Face.}
		\begin{tabular}{l|c|c}
			\hline  Method                                            & Accuracy (\%)             & Time (ms) \\ \hline
			SRC (5 per person) \cite{wright2009robust}                & 66.50                     & 17.76 \\			
			LLC (30 local bases) \cite{wang2010locality}              & 69.50                     & - \\
			DL-COPAR \cite{wang2014classification}                    & 83.29 $\pm$ 1.23          & 36.49 \\
	        FDDL \cite{yang2014sparse}                                & 85.97 $\pm$ 1.23          & 50.03 \\	
	        DBDL \cite{akhtar2016discriminative}                      & 86.15 $\pm$ 1.19          & 1.20 \\
	        K-SVD (5 per person) \cite{aharon2006k}                   & 86.50                     & - \\
	        JBDC \cite{akhtarjoint}                                   & 87.17 $\pm$ 0.99          & 1.18 \\
			LLC (70 local bases) \cite{wang2010locality}              & 88.70                     & - \\
			D-KSVD (5 per person) \cite{zhang2010discriminative}      & 88.80                     & - \\
			LC-KSVD1 (5-5) \cite{jiang2011learning}                   & 92.50                     & 0.541 \\
			LC-KSVD2 (5-5) \cite{jiang2011learning}                   & 93.70                     & \textbf{0.479} \\
			MBAP~\cite{bao2016dictionary}                                & 93.88                     & - \\ 
			MDDL-2~\cite{song2019multi}                                  & 94.9                       & - \\
			MDDL-3~\cite{song2019multi}                                  & 95.0                       & - \\			
			RRC\_L$_1$\cite{yang2013regularized}                      & 96.30                     & - \\				
			ADDL (5 items, 20 labels) \cite{zhang2017jointly}         & 97.00                     & - \\
			SRC (all training samples) \cite{wright2009robust}        & 97.50                     & 83.79 \\
			LC-KSVD2 (A-5) \cite{jiang2011learning}                   & 97.80                     & - \\
			LGII \cite{Soodeh15}                                                & 99.00                     & - \\ \hline
			PCANet-1 \cite{chan2015pcanet}                            & 98.00                     & - \\
			PCANet-2 \cite{chan2015pcanet}                            & 99.50                     & - \\ \hline
			DDLCN-2 (1-1)                                               & 99.56 $\pm$ 0.21          & 0.73 \\
			DDLCN-2 (5-5)                                               & 99.84 $\pm$ 0.36          & 1.26 \\
			DDLCN-2 (A-5)                                               & 99.87 $\pm$ 0.19 & 1.63 \\ 
			DDLCN-3 (A-5)                                               & \textbf{99.92 $\pm$ 0.15} & 1.82 \\ \hline
		\end{tabular}
	\label{table_AR}
	\vspace{-0.4cm}
\end{table}

\noindent \textbf{Extended YaleB.} 
We adopt both state-of-the-art  dictionary learning methods and deep learning methods as our baselines.
Specifically, we compare the proposed DDLCN with dictionary learning methods such as  D-KSVD~\cite{zhang2010discriminative}, LC-KSVD~\cite{jiang2011learning} and  MDDL~\cite{song2019multi}.
We also compare DDLCN with deep learning models including PCANet~\cite{chan2015pcanet} and VAE+GAN \cite{mathieu2016disentangling}.
Results are shown in the second column of Table~\ref{table:extendedYaleB}, we can see that  the proposed DDLCN achieves better results than all the dictionary learning methods.
Moreover, we observe that the proposed DDLCN achieves slightly worse results than deep learning based models such as~\cite{chan2015pcanet} and~\cite{yang2013regularized}.
However, the proposed DDLCN still outperforms the PCANet-1 version \cite{chan2015pcanet} (97.77\%), which proves the effectiveness of the proposed deep dictionary learning framework.

Moreover, we follow the evaluation metric of~\cite{jiang2011learning} and conduct another experiment with the bad images excluded to further validate the effectiveness of the proposed method.
The comparison results are shown in Table~\ref{table:extendedYaleB} (the third column).
We observe that our DDLCN obtains the best classification accuracy compared to the other methods.
Lastly, we list the computation time of predicting a single image during the testing stage.
Results compare with SRC~\cite{wright2009robust}, LC-KSVD~\cite{jiang2011learning}, DBDL~\cite{akhtar2016discriminative} and
JBDC~\cite{akhtarjoint}  are shown in Table~\ref{table:extendedYaleB} (the fourth column).
We can see that the proposed DDCLN costs remarkably less time than other baselines.

\noindent \textbf{AR Face.} 
We adopt several advanced methods such as LC-KSVD~\cite{jiang2011learning} and MDDL~\cite{song2019multi}  as our baselines.
The comparison results are reported in Table~\ref{table_AR}.
We can see that the proposed DDLCN achieves better results than other baselines including~\cite{chan2015pcanet} and~\cite{yang2013regularized} when only using a 1-1 strategy, which strongly validates the effectiveness of the proposed deep dictionary learning framework. 

Moreover, we show the computation time of predicting an image during the testing stage to simultaneously highlight the efficiency of the proposed DDLCN.
Results are shown in the third column of Table~\ref{table_AR}, we see that DDLCN only spends more time than~LC-KSVD, but costs significantly less time than SRC, DL-COPAR, JBDC, DBDL and FDDL.
It means that the proposed DDLCN is not only a robust method but also an efficient solution for image nonrecognition tasks.

\begin{table*}[!htbp] \small
	\centering
	\caption{Classification accuracy (\%) on Caltech 101.}
	\begin{tabular}{l|c|c|c|c|c|c}
		\hline  Number of Train. Samp.                           & 5        & 10        & 15          & 20        & 25         & 30  \\ \hline
		KC \cite{van2008kernel}                                    & -               & -               & -                & -               & -               & 64.14 $\pm$ 1.18  \\
		Griffin \cite{griffin2007caltech}                          & 44.20           & 54.50           & 59.00            & 63.30           & 65.80           & 67.60  \\
		SRC \cite{wright2009robust}                                & 48.80           & 60.10           & 64.90            & 67.70           & 69.20           & 70.70  \\
		D-KSVD \cite{zhang2010discriminative}                      & 49.60           & 59.50           & 65.10            & 68.60           & 71.10           & 73.00  \\
		K-SVD \cite{aharon2006k}                                   & 49.80           & 59.80           & 65.20            & 68.70           & 71.00           & 73.20  \\
		ScSPM \cite{yang2009linear}                                & -               & -               & 67.00 $\pm$ 0.45 & -               & -               & 73.20 $\pm$ 0.54  \\
		LC-KSVD1 \cite{jiang2011learning}                          & 53.50           & 61.90           & 66.80            & 70.30           & 72.10           & 73.40  \\
		LLC \cite{wang2010locality}                                & 51.15           & 59.77           & 65.43            & 67.74           & 70.16           & 73.44  \\
		LC-KSVD2 \cite{jiang2011learning}                          & 54.00           & 63.10           & 67.70            & 70.50           & 72.30           & 73.60  \\
		MBAP~\cite{bao2016dictionary}   &  54.8           & 63.6            & 68.3             & 72.2            & 72.7            & 73.9 \\ 
		EasyDL \cite{quan2016sparse}                               & -               & -               & 68.40            & -               & -               & - \\
		MDDL-2~\cite{song2019multi}                              & -               & -               & -           & -               & -               & 77.4 \\
		MDDL-3~\cite{song2019multi}                              & -               & -               & -           & -               & -               & 77.6 \\ \hline
		Deep Convolutional Learning \cite{pu2015generative}        & -               & -               & \textbf{75.24}            & -               & -               & \textbf{82.78} \\ \hline
		DDLCN-2 (1-1)                                                & 48.39 $\pm$ 1.29           & 56.77 $\pm$ 0.62           & 60.90 $\pm$ 0.61   & 64.34 $\pm$ 0.55           & 66.96 $\pm$ 0.40         & 68.75 $\pm$ 0.37 \\
		DDLCN-2 (31-30)                                              &58.46 $\pm$ 0.84   & 67.26 $\pm$ 0.98  & 72.40 $\pm$ 0.59   & 76.53 $\pm$ 0.51  & 78.90 $\pm$ 0.42& 80.16 $\pm$ 0.36 \\
		DDLCN-3 (31-30)                                              &\textbf{60.21 $\pm$ 0.62}   & \textbf{70.12 $\pm$ 0.65}  & 74.62 $\pm$ 0.48   & \textbf{77.92 $\pm$ 0.39}  & \textbf{80.13 $\pm$ 0.26}& 81.98 $\pm$ 0.27 \\ \hline
	\end{tabular}
	\label{table_caltech101_t}
	\vspace{-0.4cm}
\end{table*}

\begin{figure}[!tbp] \small
	\centering
	\setcounter{subfigure}{0}
	\subfigure{\includegraphics[height=1.32in]{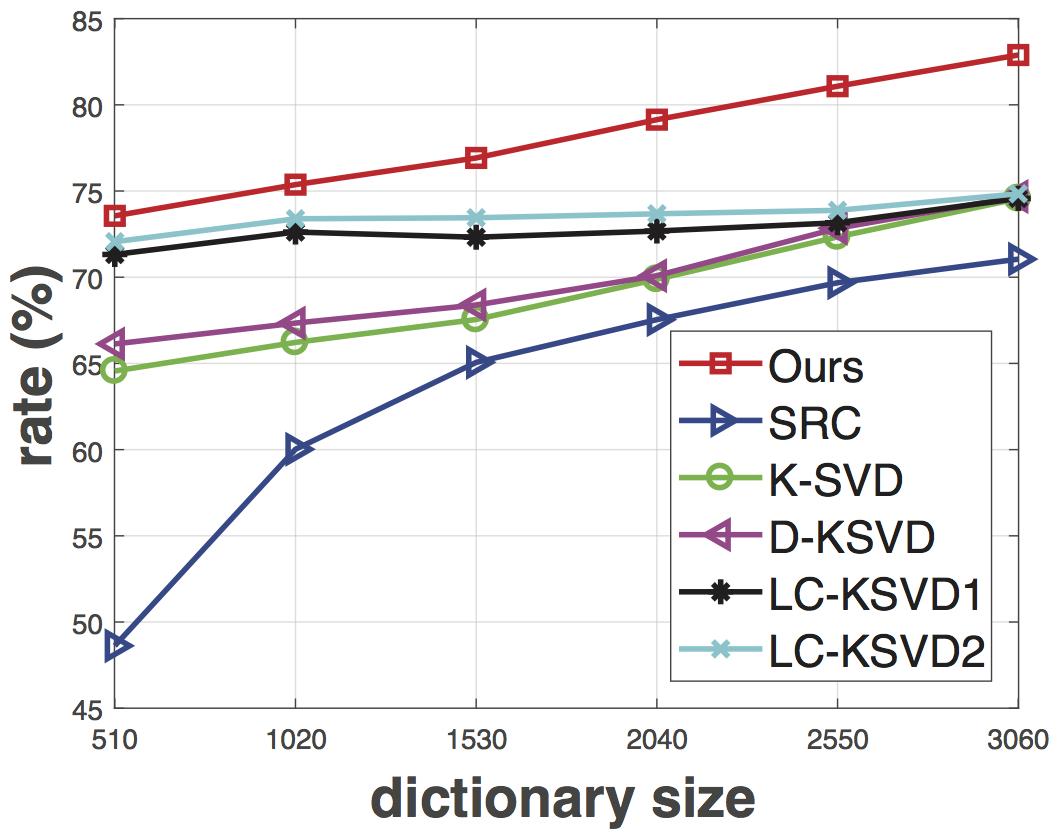}}
	\subfigure{\includegraphics[height=1.32in]{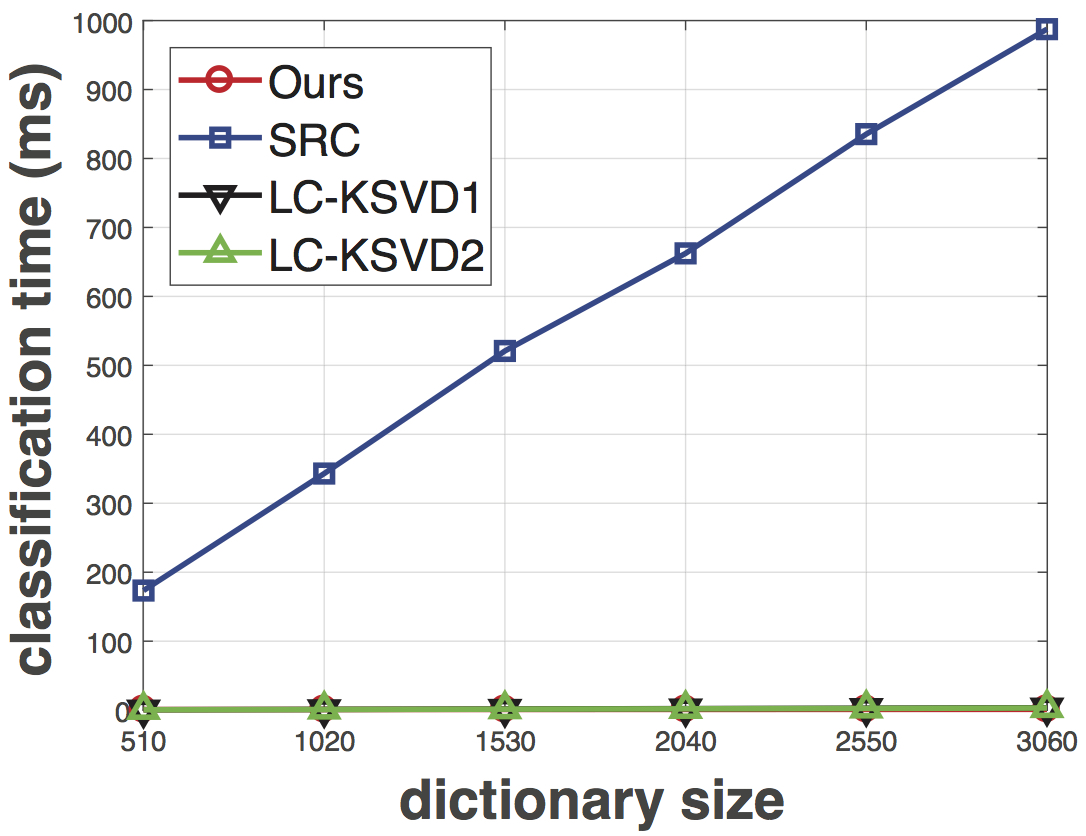}}\\
	\caption{Performance comparisons on Caltech 101. (\textit{left}) Performance comparisons with varying dictionary size. (\textit{right}) Computation time (ms) for classifying an image during testing.}
	\label{Fig:caltech101_dic_and_time}
	\vspace{-0.4cm}
\end{figure}

\begin{figure}[!tbp] \small
	\centering
	\centering
	\setcounter{subfigure}{0}
	\subfigure[sea horse, acc: 29.63\%]{\includegraphics  [width=1\linewidth]{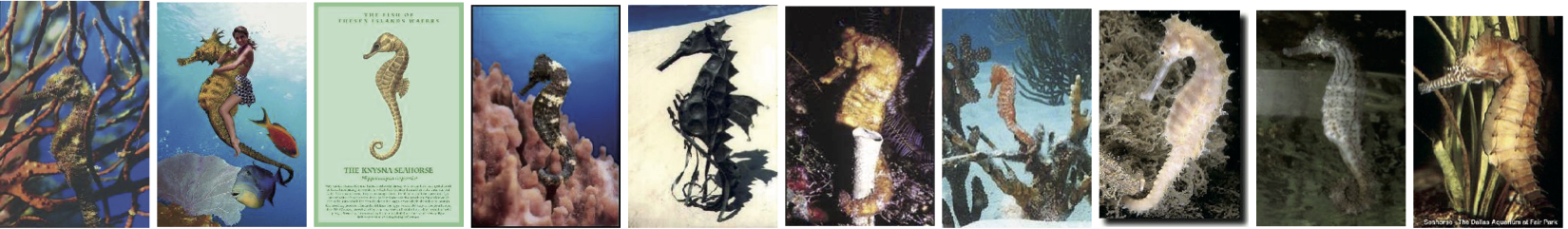}}\\
	\subfigure[cougar body, acc: 23.53\%]{\includegraphics[width=1\linewidth]{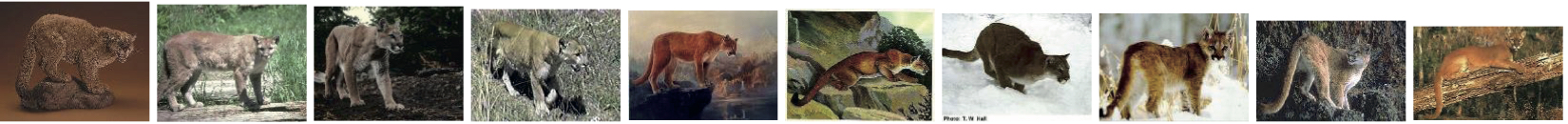}}\\
	\subfigure[octopus, acc: 20.00\%]{\includegraphics    [width=1\linewidth]{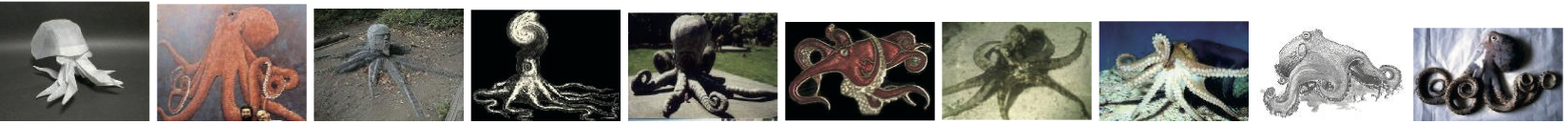}}\\
	\subfigure[ant, acc: 16.67\%]{\includegraphics        [width=1\linewidth]{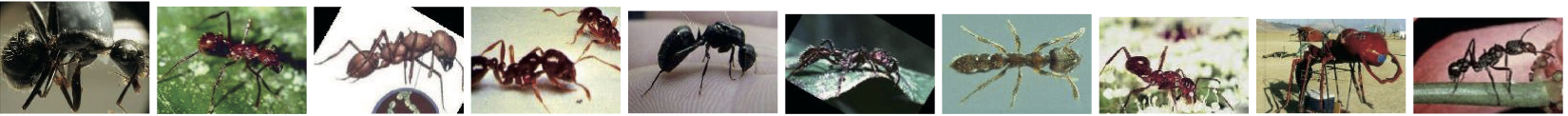}}\\
	\subfigure[beaver, acc: 6.25\%]{\includegraphics      [width=1\linewidth]{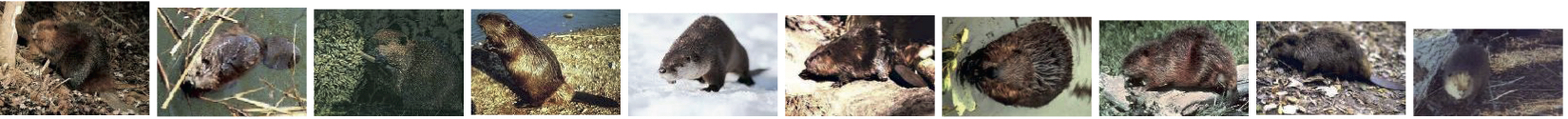}}\\
	\caption{The five categories with the lowest classification accuracy on Caltech 101.}
	\label{fig: caltech101_acc_worst}
	\vspace{-0.4cm}
\end{figure}

\noindent \textbf{Caltech 101.} 
We follow the standard experimental settings and randomly select 5, 10, 15, 20, 25, and 30 images per category for training and the remaining for testing.
In Table~\ref{table_caltech101_t}, we observe that the proposed method achieves better classification performance in all the situations except when using 15 and 30 training samples.
The classification accuracy of DDLCN-2 are slightly worse than~\cite{pu2015generative}, while \cite{pu2015generative} adopts a three-layer model.
In this case,~\cite{pu2015generative} achieves a higher recognition accuracy since more stacked layers can enrich the `level' of features as indicated in \cite{he2016deep}.
We also note that such CNN-based methods are supervised while our DDLCN is an unsupervised method.
Moreover, DDLCN-3 further achieves competitive results compared with~\cite{pu2015generative}.
Currently, our method only uses one hand-crafted feature, one can always use more powerful image-level features such as CNN-based features for further improving the performance.
Besides, we observe that DDLCN outperforms other methods when using only a few training samples such as 5 and 10 for each category, which is of great benefit when the training data is limited.

Next, we follow~\cite{jiang2011learning} and evaluate the performance of our method with different dictionary sizes.
Specifically, we set $K {=} [510, 1020, 1530, 2040, 2550, 3060]$, respectively.
Fig.~\ref{Fig:caltech101_dic_and_time}(\textit{left}) shows that the proposed DDLCN achieves better classification accuracy than state-of-the-art dictionary learning methods such as K-SVD, D-KSVD, SRC and LC-KSVD.
In Fig.~\ref{Fig:caltech101_dic_and_time}(\textit{right}), we show the computational speed of classifying one test image using different dictionary sizes. 
We can see that our method is remarkably faster than SRC, and is very close to LC-KSVD.
Lastly, when using 30 images per category for training, our method achieves 100\% classification accuracy on 9 classes, i.e., accordion, car side, garfield, inline skate, metronome, minaret, okapi, snoopy and trilobite.
We also present the five categories with the lowest classification accuracy in Fig. \ref{fig: caltech101_acc_worst}, which are sea horse, cougar body, octopus, ant and beaver.
We can observe that the five categories are all moving animals with significant differences in shape, pose, color and background.

\begin{table}[!t] \small
	\centering
	\caption{Classification accuracy (\%) on Caltech 256.}
	\resizebox{\linewidth}{!}{
		\begin{tabular}{l|c|c|c|c}
			\hline  Num. of Train. Samp.            & 15         & 30          & 45          & 60  \\ \hline
			
			KC \cite{van2008kernel}                 & -                & 27.17 $\pm$ 0.46 & -                & - \\
			LLC \cite{wang2010locality}             & 25.61            & 30.43            & -                & - \\
			K-SVD \cite{aharon2006k}                & 25.33            & 30.62            & -                & - \\
			D-KSVD \cite{zhang2010discriminative}   & 27.79            & 32.67            & -                & - \\
			LC-KSVD1 \cite{jiang2011learning}       & 28.10            & 32.95            & -                & - \\
			SRC \cite{wright2009robust}             & 27.86            & 33.33            & -                & - \\			
			Griffin \cite{griffin2007caltech}       & 28.30            & 34.10 $\pm$ 0.20 & -                & - \\
			LC-KSVD2 \cite{jiang2011learning}       & 28.90            & 34.32            & -                & - \\	
			ScSPM \cite{yang2009linear}             & 27.73 $\pm$ 0.51 & 34.02 $\pm$ 0.35 & 37.46 $\pm$ 0.55 & 40.14 $\pm$ 0.91 \\
			NDL \cite{hu2018nonlinear}              & 29.30 $\pm$ 0.29 & 36.80 $\pm$ 0.45 & -                & - \\
			SNDL \cite{hu2018nonlinear}             & 31.10 $\pm$ 0.35 & 38.25 $\pm$ 0.43 & -                & - \\
			MLCW \cite{fanello2014ask}              & 34.10            & 39.90            & 42.40            & 45.60 \\
			LP-$\beta$ \cite{gehler2009feature}      & -                & 45.8             & -                & - \\	
			M-HMP \cite{bo2013multipath}            & 42.7             & 50.7             & 54.8             & 58.0 \\ \hline
			Convolutional Networks \cite{zeiler2014visualizing}    & - & -                & -                & 74.2 $\pm$ 0.3 \\
			VGG19  \cite{simon2015neural}           & -                & -                & -                & \textbf{84.10} \\ \hline
			DDLCN-2 (1-1)   & 26.30 $\pm$ 0.40             & 31.45 $\pm$ 0.21           & 34.69 $\pm$ 0.31          & 37.76 $\pm$ 0.25        \\
			DDLCN-2 (15-15) & 35.06 $\pm$ 0.26  & 41.26 $\pm$ 0.22  & 44.17 $\pm$ 0.35 & 47.48 $\pm$ 0.26 \\
			DDLCN-2 (30-30) & 45.25 $\pm$ 0.31  & 51.64 $\pm$ 0.51  & 55.11 $\pm$ 0.26 & 59.66 $\pm$ 0.45 \\
			DDLCN-3 (30-30) & \textbf{47.65 $\pm$ 0.22}  & \textbf{54.28 $\pm$ 0.42}  & \textbf{57.89 $\pm$ 0.32} & 62.42 $\pm$ 0.34 \\ \hline
	\end{tabular}}
	\label{table_Caltech256}
	\vspace{-0.4cm}
\end{table}

\noindent \textbf{Caltech 256.}
We conduct extensive experiments using 15, 30, 45 and 60 training images per class and compare with state-of-the-art methods.
Table~\ref{table_Caltech256} shows the comparison results.
We can see that the proposed DDLCN outperforms existing leading dictionary-based methods such as~K-SVD, D-KSVD, LC-KSVD and LLC, which  significantly validates the advantages of the proposed DDLCN.

\begin{figure}[!t]  \small
	\centering
	\centering
	\setcounter{subfigure}{0}
	\subfigure[kayak,      acc: 2.33\%]{\includegraphics[width=0.49\linewidth]{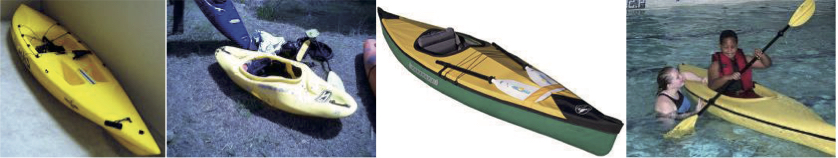}}
	\subfigure[skateboard, acc: 2.33\%]{\includegraphics[width=0.49\linewidth]{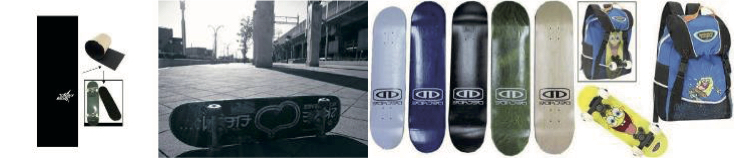}} \\
	\subfigure[canoe,      acc: 2.27\%]{\includegraphics[width=0.49\linewidth]{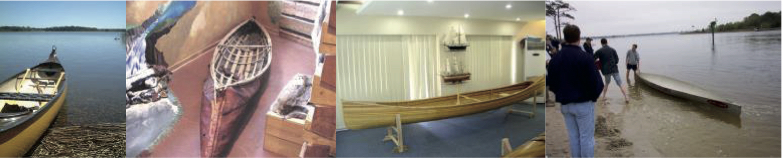}}
	\subfigure[hot dog,    acc: 0.00\%]{\includegraphics[width=0.49\linewidth]{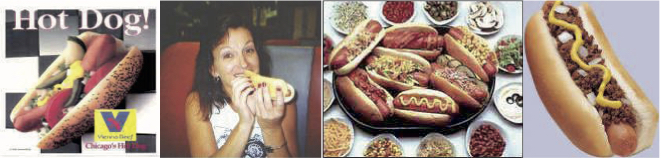}} \\
	\subfigure[rifle,      acc: 0.00\%]{\includegraphics[width=0.49\linewidth]{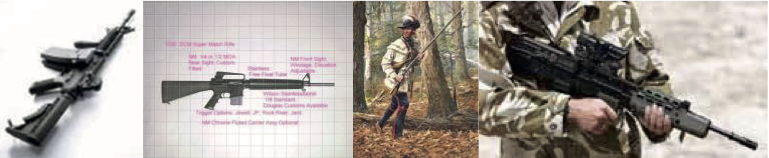}}
	\subfigure[soda can,   acc: 0.00\%]{\includegraphics[width=0.49\linewidth]{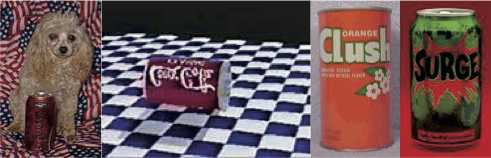}} \\
	\caption{The six categories with the lowest classification accuracy on Caltech 256.}
	\label{Fig: caltech256_acc_worst}
	\vspace{-0.4cm}
\end{figure}

Moreover, we observe that the proposed method achieves slightly worse results than both VGGNet~\cite{simon2015neural} and convolutional network~\cite{zeiler2014visualizing} when using 60 training samples.
However, 1) \cite{simon2015neural} uses a very deep convolutional network, i.e., VGG19~\cite{simonyan2014very}, which consists of 16 convolutional layers and 3 fully connected layers. 
2) Both \cite{simon2015neural}  and \cite{zeiler2014visualizing} have limited practical applicability than our DDLCN since they rely on careful hyper-parameter selection.
3) The proposed approach has much fewer hyper-parameters that need to be tuned.
4) Compared with~\cite{simon2015neural} and~\cite{zeiler2014visualizing}, the feature learner and encoder of the proposed DDLCN are fixed after extracting features, and only the linear SVM classifier on top is needed to update during training. 
Thus, the training of DDLCN is offline and its testing is pretty fast.
All these represent the big advantages of the proposed DDLCN. 
Lastly, we show the six categories with the lowest classification accuracy in Fig. \ref{Fig: caltech256_acc_worst} when using 60 training images per category, which are kayak, skateboard, canoe, hot dog, rifle and soda can.

\noindent \textbf{MNIST.} The results on MNIST are shown in Table~\ref{table_mnist}.
We observe that the proposed DDLCN-2 consistently outperforms all the baselines except~\cite{pu2015generative} when using the 500-500 strategy.
Pu et al. \cite{pu2015generative}~employs a three layers model to achieve a slightly better result (+1.03\%) than us.
The reason is that such CNN-based methods are jointly optimized between forward and backward propagation, while our DDLCN has no end-to-end tuning.
Therefore, the training of our DDLCN is more efficient than such CNN-based models.

\begin{figure}[!t] \small
	\centering
	\centering
	\setcounter{subfigure}{0}
	\subfigure{\includegraphics[width=0.49\linewidth]{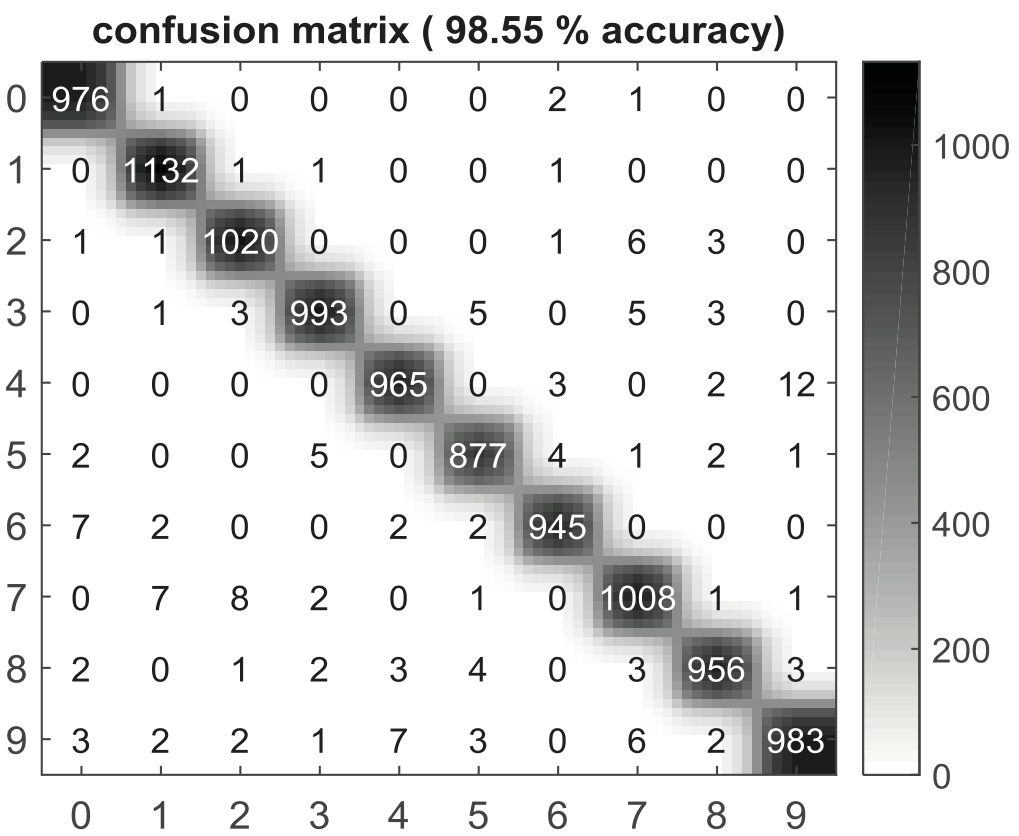}}
	\subfigure{\includegraphics[width=0.49\linewidth]{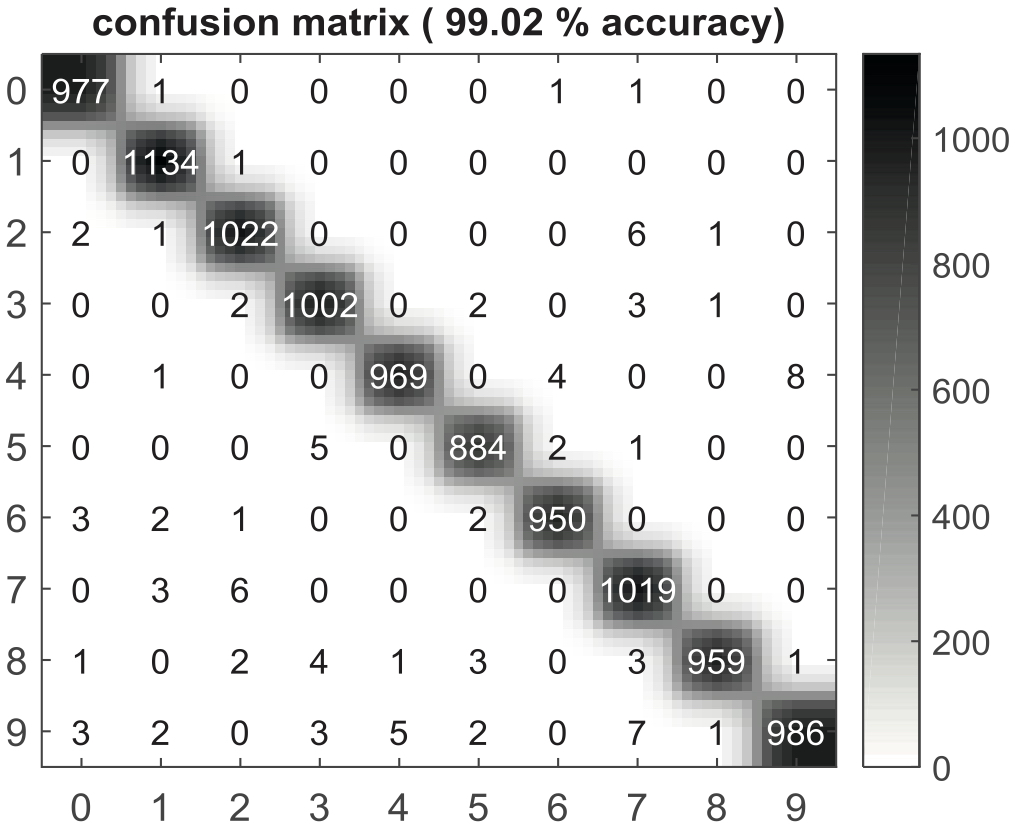}}\\
	\caption{Confusion matrix on MNIST.}
	\label{Fig:mnist_con_500_500}
	\vspace{-0.4cm}
\end{figure}

Moreover, when setting $p{=}1$ and $q{=}2$, the proposed model achieves 96.56\% classification accuracy, which proves again that the proposed  DDLCN can also achieve good performance when the number of training samples is limited and the size of the dictionary is small.
This advantage is of great benefit in practical applications when the training data is limited.
Also, we observe that when $p{=}100$ and $q{=}100$, the proposed method achieves   98.55\% classification accuracy. 
We also show the confusion matrix of each category in Fig.~\ref{Fig:mnist_con_500_500}(\textit{left})).
When setting $p{=}500$ and $q{=}500$, the  performance is further boosted.
Specifically, we achieve 99.02\% classification accuracy on this dataset.
Fig.~\ref{Fig:mnist_con_500_500}(\textit{right}) lists the confusion matrix of each category under this experimental setting, and we can see that the most confusing pairs are (2, 7), (4, 9) and (3, 5).

Finally, we note that the classification accuracy of our hierarchical DDLCN increases by adding more layers.
For example, the performance of DDLCN-3 achieves better results than the shallow one, i.e., DDLCN-2.
To further explore the upper bound of our method, we conduct experiments on MNIST using DDLCN-4, DDLCN-5 and DDLCN-6.
Specifically, DDLCN-4, DDLCN-5 and DDLCN-6 adopting 4, 5 and 6 the proposed compound dictionary learning and coding layers, respectively, and keeping all other details the same. 
Results are reported in Table~\ref{table_mnist}. 
We observe that as the number of proposed dictionary learning and coding layers increases, the classification accuracy also gradually improves. However, when the number of layers increases to a certain number (`6' in this case), the classification performance saturates, which can be also observed in deep CNN models \cite{simon2015neural,zeiler2014visualizing}.

\begin{table}[!t] \small
	\centering
	\caption{Classification accuracy (\%) on MNIST.}
	\begin{tabular}{l|c} \hline
		Method                                                    & Accuracy \\ \hline
		Deep Representation Learning \cite{yang2015deep}          & 85.47          \\
		D-KSVD~\cite{zhang2010discriminative}                     & 90.33          \\ 
		LC-KSVD~\cite{jiang2013label}                             &  92.58          \\
		SRC~\cite{wright2009robust}                               & 95.69          \\
		DCN \cite{lin2010deep}                                    & 98.15          \\ 
		TLCC \cite{xiao2015two} & 98.57 \\ \hline
		Embed CNN \cite{weston2012deep}                           & 98.50          \\
		Convolutional Clustering \cite{dundar2015convolutional}   & 98.60          \\
		Deep Convolutional Learning \cite{pu2015generative}       & \textbf{99.58} \\ \hline
		DDLCN-2 (1-2)       & 96.56   \\
		DDLCN-2 (100-100)   & 98.55   \\
		DDLCN-2 (500-500)   & 99.02    \\
		DDLCN-3 (500-500)   & 99.35              \\ 
		DDLCN-4 (500-500)   & 99.47   \\ 
		DDLCN-5 (500-500)   & 99.54 \\ 
		DDLCN-6 (500-500)   & 99.57 \\ \hline
	\end{tabular}
	\vspace{-0.4cm}
	\label{table_mnist}
\end{table}

\section{Conclusion}
\label{conclusions}
In this paper, we aim to  improve the deep representation capability of  dictionary learning. 
To this end, we propose a novel deep dictionary learning method, i.e., DDLCN,  to learn multi-layer deep dictionaries, which combines the advantages of both deep learning and dictionary learning approaches, and achieves promising performance. 
Moreover, we propose a novel dictionary learning and coding layer and use it to substitute traditional convolutional layers in CNNs.
Extensive experimental results on five public datasets with limited training data show that our DDLCN outperforms leading dictionary learning methods and achieves competitive results compared with state-of-the-art CNN-based models.

\noindent \textbf{Acknowledgments.}
This work is partially supported by National Natural Science Foundation of China (No.U1613209, 61673030), National Natural Science Foundation of Shenzhen (No.JCYJ20190808182209321), and by the Italy-China collaboration project TALENT.

\footnotesize
\bibliographystyle{IEEEtran}
\bibliography{ref}

\end{document}